%% file: main-9013-Wolfson.tex
\documentclass[11pt,a4paper]{article}
\usepackage{times,latexsym}
\usepackage{url}
\usepackage[T1]{fontenc}
\usepackage{mathtools}

\usepackage{multirow}
\usepackage{wrapfig,lipsum}
\usepackage{arydshln}
\usepackage{booktabs}
\usepackage{authblk} 
\usepackage{array}
\usepackage{soul}

\makeatletter

\newcommand\email[2][]%
   {\newaffiltrue\let\AB@blk@and\AB@pand
      \if\relax#1\relax\def\AB@note{\AB@thenote}\else\def\AB@note{\relax}%
        \setcounter{Maxaffil}{0}\fi
      \begingroup
        \let\protect\@unexpandable@protect
        \def\thanks{\protect\thanks}\def\footnote{\protect\footnote}%
        \@temptokena=\expandafter{\AB@authors}%
        {\def\\{\protect\\\protect\Affilfont}\xdef\AB@temp{#2}}%
         \xdef\AB@authors{\the\@temptokena\AB@las\AB@au@str
         \protect\\[\affilsep]\protect\Affilfont\AB@temp}%
         \gdef\AB@las{}\gdef\AB@au@str{}%
        {\def\\{, \ignorespaces}\xdef\AB@temp{#2}}%
        \@temptokena=\expandafter{\AB@affillist}%
        \xdef\AB@affillist{\the\@temptokena \AB@affilsep
          \AB@affilnote{}\protect\Affilfont\AB@temp}%
      \endgroup
       \let\AB@affilsep\AB@affilsepx
}
\makeatother

   \usepackage[acceptedWithA]{tacl2021v1}
\usepackage{xspace,mfirstuc,tabulary}

\usepackage[shortlabels]{enumitem}

\newcommand\blfootnote[1]{%
  \begingroup
  \renewcommand\thefootnote{}\footnote{#1}%
  \addtocounter{footnote}{-1}%
  \endgroup
}

\usepackage{todonotes}

\newif\iftaclinstructions
\taclinstructionsfalse 
\iftaclinstructions
\renewcommand{\confidential}{}
\renewcommand{\anonsubtext}{(No author info supplied here, for consistency with
TACL-submission anonymization requirements)}
\newcommand{\instr}
\fi

\iftaclpubformat 

\else

\fi

\usepackage{arydshln}

\newcommand\dataset{\textsc{MoNaCo}}
\newcommand\datasize{1,315}
\newcommand\datasubqsize{90K}
\newcommand\datalistqsize{8,549}

\newcommand\hotpot{\textsc{HotpotQA}}
\newcommand\fanout{\textsc{FanOutQA}}
\newcommand\drop{\textsc{DROP}}
\newcommand\qampari{\textsc{QAMPARI}}
\newcommand\quest{\textsc{QUEST}}
\newcommand\musique{\textsc{Musique}}
\newcommand\mintaka{\textsc{Mintaka}}
\newcommand\naturlquestions{\textsc{Nat. Questions}}

\newcommand\gptfive{\textsc{GPT-5}}

\newcommand\gptfouro{\textsc{GPT-4o}}
\newcommand\gptfourone{\textsc{GPT-4.1}}
\newcommand\gptfourturbo{\textsc{GPT-4 Turbo}}
\newcommand\llamathree{\textsc{Llama 3-70B}}
\newcommand\llamathreeone{\textsc{Llama 3.1-405B}}
\newcommand\othree{\textsc{o3}}
\newcommand\qwentwo{\textsc{Qwen 2-72B}}
\newcommand\qwentwofive{\textsc{Qwen 2.5-72B}}
\newcommand\othreemini{\textsc{o3-mini}}
\newcommand\ofourmini{\textsc{o4-mini}}
\newcommand\rone{\textsc{Deepseek-R1}}
\newcommand\vthree{\textsc{Deepseek-V3}}
\newcommand\geminipro{\textsc{Gemini 2.5-Pro}}
\newcommand\geminiflash{\textsc{Gemini 2.5-Flash}}
\newcommand\claudefour{\textsc{Claude 4-Opus}}

\title{\dataset{}: More Natural and Complex Questions for Reasoning\\ Across Dozens of Documents}

\author[1]{\bf Tomer Wolfson}
\author[2]{\bf Harsh Trivedi}
\author[3]{\bf Mor Geva}
\author[2,4]{\bf Yoav Goldberg}
\author[1,5]{\\\bf Dan Roth}
\author[*]{\bf Tushar Khot}
\author[2]{\bf Ashish Sabharwal}
\author[4]{\bf Reut Tsarfaty}
{
\email{\vspace{-0.7cm}}
\makeatletter
\renewcommand\AB@affilsepx{\quad \protect\Affilfont}
\makeatother
\affil[1]{University of Pennsylvania}
\affil[2]{Allen Institute for AI}
\affil[3]{Tel Aviv University}
}

\email{\vspace{0.05cm}}

{
\makeatletter
\renewcommand\AB@affilsepx{\quad \protect\Affilfont}
\makeatother
\affil[4]{Bar-Ilan University}
\affil[5]{Oracle AI}
}

\email{\vspace{-0.3cm}}

\email{ \texttt{wolfsont@seas.upenn.edu}}

\date{}

\begin{document}

\maketitle
\input{00_abstract}

\iftaclpubformat

\input{01_introduction}

\input{03_executor}

\input{04_dataset}
\input{05_data_analysis}
\input{06_llm_models}

\input{07_related_work}

\input{08_conclusion}
\input{09_limitations}
\section*{Acknowledgments}
We would like to thank Ron Yachini of AI2 Israel for his help and support. We would also like to thank Ori Yoran, Chaitanya Malaviya, Ofir Press, Roee Aharoni, Avi Caciularu, Allen Chang, Peter Clark, Jonathan Berant and Daniel Deutch for their feedback and insightful comments. Lastly, we wish to express our gratitude to our action editor and reviewers for their time and effort. This work was done while the first author was an intern at the Allen Institute for AI. In addition, this project was partially funded by ONR Contract N00014-23-1-2364. The last author was funded by The Israel Science Foundation grant 670/23, for which we are grateful.

\bibliography{tacl2021}
\bibliographystyle{acl_natbib}


\onecolumn

\input{00_appendix}

\end{document}

%% file: 00_abstract.tex

\begin{abstract}

Automated agents, powered by Large language models (LLMs), are emerging as the go-to tool for querying information.
However, evaluation benchmarks for LLM agents rarely feature natural questions that are both information-seeking and genuinely time-consuming for humans.
To address this gap we introduce \dataset{}, a benchmark of \datasize{} natural and time-consuming questions that require dozens, and at times hundreds, of intermediate steps to solve --- far more than any existing QA benchmark. 
To build \dataset{},
we developed a decomposed annotation pipeline to elicit and manually answer real-world time-consuming questions at scale. Frontier LLMs evaluated on \dataset{} achieve at most 61.2\% F1, hampered by low recall and hallucinations. 
Our results underscore the limitations of LLM-powered agents in handling the complexity and sheer breadth of real-world information-seeking tasks --- with \dataset{} providing an effective resource for tracking such progress.  
The \dataset{} benchmark, codebase, prompts and models predictions are all publicly available at: \url{https://tomerwolgithub.github.io/monaco}.



\end{abstract}

%% file: 01_introduction.tex
\section{Introduction}
\label{sec:intro}

\begin{figure}[t]\setlength{\belowcaptionskip}{-8pt}
  \includegraphics[clip, width=0.45\textwidth]{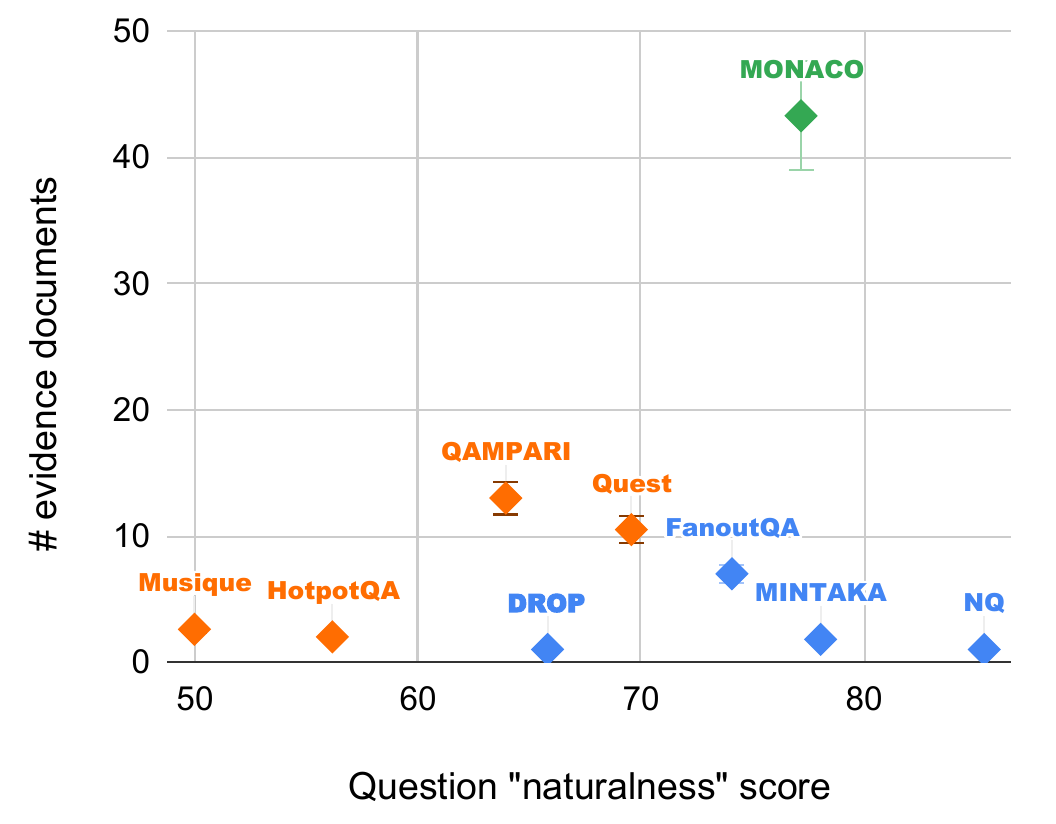}
  \centering
  \vspace*{-0.4cm}
  \caption{\dataset{} (in green) and 8 QA benchmarks, plotted based on their ``naturalness'' (judged by humans) and their number of evidence documents. Natural QA benchmarks are in blue, artificial ones in orange. \dataset{} targets the top-right region: natural questions that require dozens of evidence documents.}
  \label{fig:monaco_natural_complexity}
\end{figure}

\blfootnote{* Work done while the author was at Allen AI.}Large language models (LLMs), and the ``agentic'' systems built around them, are becoming increasingly ingrained into how people seek and query information \cite{Guo2024LargeLM,OpenAIDeepRS}.
It is therefore important to understand
how well do LLM-powered agents fare in answering real-world questions that require combining information from dozens or even hundreds of documents. For example, a political scientist may want to know whether ``\textit{In European countries, are left-wing political parties  more likely to be headed by women than right-wing ones?}'' a question which, unless answered verbatim in some source, is extremely time-consuming.
In fact, answering this question entails reviewing all current political parties in each European country, extracting their political affiliation and leader's gender, and then to reason over all these intermediate facts, combining results from 719 distinct pages (see Table~\ref{tab:monaco_examples}).

\begin{table*}[t]
\centering
\scriptsize
  \begin{tabular}{p{0.095\linewidth}p{0.68\linewidth}cc}
    \toprule
    \bf Domain & \bf Example & \bf\# Pages & \bf\# Steps\\ 
    \midrule
    \multirow{1}{*}{Politics} & In European countries, are current left-wing political parties more likely to be headed by women than right-wing ones?  & 719 & 16 
  \\\midrule
    \multirow{1}{*}{History} & What has been the highest percentage of parliament seats held by monarchist parties during the time of the Third French Republic?  & 49 & 10 
    \\\midrule
    \multirow{1}{*}{Demographics} 
    & Which Nobel Prize category has the fewest number of Asian-born recipients?  & 871 & 6 \\\midrule
    \multirow{1}{*}{Sports} 
    & What was the youngest team in the NBA in 2021?  & 550 & 5 \\\midrule
    \multirow{1}{*}{Culinary} 
    & I want to cook a traditional Bulgarian meal for my girlfriend, that is allergic to eggs and dairy, which dishes would be okay?  & 42 & 3 \\\midrule
    \multirow{1}{*}{Art} & Which museums house the most famous paintings by each of the leading French Impressionists? & 58 & 3 
    \\\midrule
    \multirow{1}{*}{Literature} 
    & Which books by Gabriel Garcia Marquez are based on real historical events?  & 21 & 3 \\\midrule
    \multirow{1}{*}{Music} 
    & What percentage of musicians with Billboard Year-end Hot 100 singles were born outside the United States in 2020, 2010 and 2000? & 449 & 22
    \\\midrule
    \multirow{1}{*}{Film \& TV} & What are the 3 most common professions among the fathers of Oscar winners for best actress?   & 75 & 5
  \\\midrule
    \multirow{1}{*}{Pop Culture} & What are the most common names for girls in the UK that are not originally Biblical?  & 21 & 3 
\\\midrule
    \multirow{1}{*}{Other} 
    & What percentage of US supreme court justices throughout history did not attend an Ivy League school for their postgraduate degree?  & 59 & 7 \\
  \bottomrule
\end{tabular}
\caption{Questions from \dataset{}, along with their number of unique Wikipedia pages containing the necessary evidence (\# Pages), and the number of decomposition steps required to answer each question (\# Steps).}
  \label{tab:monaco_examples}
\end{table*}

While LLM-powered agents hold great promise in solving realistic time-consuming tasks, such questions are not well represented in contemporary QA benchmarks \cite{ wei2024measuringshortformfactualitylarge, wei2025browsecompsimplechallengingbenchmark}.
Building a benchmark that contains challenging questions which are \emph{also} natural is no easy feat.  QA benchmarks that focus on {\em natural} questions typically contain simple questions, answerable using a single passage of text \cite{Abujabal2018ComQAAC, kwiatkowski-etal-2019-natural}. Collecting questions that are  realistic and {\em complex} has largely been reserved for domain experts, which incur high annotation costs and often result in benchmarks that are small or have questions that are not particularity time-consuming, involving only a handful of documents \cite{malaviya-etal-2024-expertqa,yoran2024assistantbenchwebagentssolve}. 
As it is non-trivial to collect complex natural questions ``in the wild'', researchers have instead focused on creating such benchmarks artificially \cite{trivedi-etal-2022-musique, li-etal-2024-deceptive, wei2025browsecompsimplechallengingbenchmark}. However, machine-based approaches often result in contrived questions that do not reflect real-world users' needs (x-axis of Figure~\ref{fig:monaco_natural_complexity}).

To address this gap we introduce \textbf{\dataset{}}, a benchmark of \textbf{Mo}re \textbf{Na}tural and much more \textbf{Co}mplex questions. This benchmark is designed to evaluate LLM-based systems on information-seeking tasks that are realistic and time-consuming, and demand planning, collecting and synthesizing many of pieces of information. 

\dataset{} contains \datasize{} challenging multi-step questions whose solution involves retrieving, filtering and aggregating dozens of intermediate facts, found in both unstructured and structured sources (paragraphs and tables). All questions in \dataset{} come with a gold-standard, manually annotated, reasoning chain that contains all of the intermediate steps, answers and supporting evidence required to solve it. Namely, each question has between 1 to 2,379 Wikipedia paragraphs and tables that are needed to solve it --- with an average of 43.3 unique pages per question.
The questions are designed to reflect the information-seeking goals of human personas (history professor, amateur chef, etc.), with a special emphasis on time-consuming tasks (Table~\ref{tab:monaco_examples}).

To construct \dataset{}, we first developed an approach for eliciting complex questions from humans. Instead of relying on pre-defined templates, we prompted workers to generate questions that would engage specific ``target personas''. This method produced questions that are not only more challenging but also perceived as more realistic, as confirmed by a user study comparing \dataset{} to 8 other natural QA benchmarks (\S\ref{sec:data_analysis}). Concretely, on the x-axis of Figure~\ref{fig:monaco_natural_complexity}, \dataset{} questions rank close to the natural yet simpler benchmarks (in blue), while being ranked much higher than all machine-generated benchmarks (in orange).

\begin{figure*}[t]\setlength{\belowcaptionskip}{-8pt}
  \centering
  \includegraphics[clip, width=0.98\textwidth]{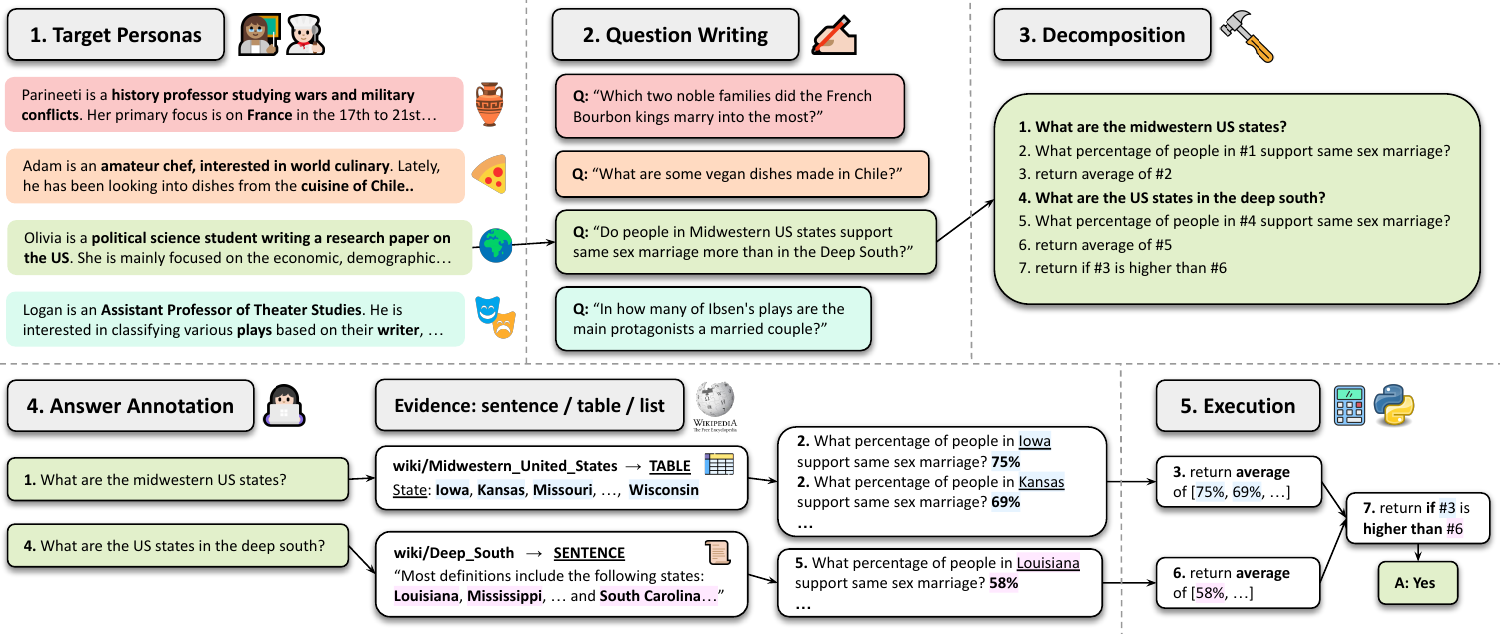}
  \vspace*{-0.4cm}
  \caption{An overview of our annotation pipeline for building the \dataset{} benchmark.}
  \label{fig:monaco_overview}
\end{figure*}

Collecting answers to \dataset{} is non-trivial as questions typically require combining information from dozens and even hundreds of documents. Therefore, we use the question decomposition method of \citet{wolfson-etal-2020-break} to implement a \emph{decomposed annotation pipeline}, breaking the process of annotating complex, time-consuming questions into multiple, simpler tasks (Figure~\ref{fig:monaco_overview}). This distributed approach  facilitates the annotation of gold answers by enabling non-expert workers to answer the simpler intermediate steps of much more challenging questions. 

As an AI benchmark, \dataset{} enables us to evaluate LLM-based systems in 4 key settings: (1) evaluating parametric knowledge and reasoning, (2) evaluating complex reasoning over long contexts when all the information is provided, (3) evaluating end-to-end retrieval-augmented QA, and (4) evaluating multi-document retrieval. In our experiments (\S\ref{sec:models_experiments}), we focus on the first three settings. First, we find that frontier LLMs, including recent AI reasoning models \cite{openai2024openaio1card, deepseekai2025deepseekr1incentivizingreasoningcapability}, struggle on \dataset{}, with the top performing \othree{} model reaching an F1 score of 61.2\% --- with only 38.7\% of the examples getting a perfect score. We observe that all models suffer from low recall and tend to hallucinate answers that involve aggregating many facts. Second, even in an Oracle retrieval setting, where all of the gold evidence is provided as context, \gptfouro{} and \llamathreeone{} struggle to perform long-context reasoning, scoring only 59\% F1.
Thirdly, retrieval-augmented generation did not help, with BM25 retrieval hurting LLMs' performance by over 12 points, exhibiting a lack of ``retrieval robustness'' \cite{yoran2024making}.

Overall, our contribution is threefold:
\begin{itemize}[topsep=0pt, itemsep=0.5pt, leftmargin=.2in, parsep=0pt]
    
    \item We design an annotation pipeline  to elicit and annotate realistic, time-consuming questions, with answers that often span across dozens and even hundreds of documents (\S\ref{sec:novel_approach}, \S\ref{sec:dataset}).

    \item We release \dataset{}, a benchmark of \datasize{} information-seeking questions that require retrieving and aggregating information from dozens of Wikipedia pages (\S\ref{sec:data_analysis}). 
    
    \item We experiment with 15 state-of-the-art LLMs and highlight the shortcomings of such models in answering \dataset{} questions (\S\ref{sec:models_experiments}).
\end{itemize}

As LLM agents near human-level performance on a variety of tasks, it becomes increasingly important to expose their limitations, especially when tackling information-seeking questions.  
To this end, \dataset{} serves as a unique and challenging testbed for evaluating, LLM-powered, agentic and ``Deep Research'' systems \cite{Huang2025DeepRA} on broad tasks that span across potentially hundreds of documents, demanding extensive factual knowledge, information retrieval and reasoning skills.

%% file: 03_executor.tex
\section{Questions, Decompositions \& Answers}
\label{sec:novel_approach}


This section describes our novel approach for collecting complex natural questions, their corresponding answers and a comprehensive set of supporting evidence (Figure~\ref{fig:monaco_overview}). 
Our first stage is \emph{question writing}, where annotators write time-consuming questions that reflect the information-seeking goals of target personas (\S\ref{sec:question_writing}). In the next stage, we annotate each question with a formal \emph{question decomposition} (\S\ref{sec:annotation_answers}). The decomposition enables us to facilitate the answer annotation by breaking down challenging questions into a series of much simpler annotation tasks. Finally, we outline our \emph{decomposition execution engine}, which queries and aggregates intermediate answers in order to derive the final answer (\S\ref{sec:annotation_execution}).

\subsection{Eliciting Complex Natural Questions}
\label{sec:question_writing}
As discussed in \S\ref{sec:intro}, collecting realistic and time-consuming questions ``in the wild'' is hard: users tend to shy away from issuing such queries to search engines \cite{kwiatkowski-etal-2019-natural, thakur2021beir}, while collecting them from user-AI interactions, most of which are private, remains challenging \cite{lin2024wildbench}.
Instead, we rely on \emph{human annotators} to write questions that reflect real-world users' information-seeking goals.

Our approach is to \emph{prompt} crowd workers to write questions that would interest a particular \emph{target persona}. 
To illustrate, Part 1 in Figure~\ref{fig:monaco_overview} presents four personas: a history professor, an amateur chef, a political scientist and a theater scholar. 
By priming workers to assume a specific persona and not to use any pre-defined question templates, we encourage them to come up with more realistic questions. In addition, the use of multiple distinct personas helps diversify our data. To encourage challenging questions, we included five randomly selected \emph{reference questions} (initially written by us and later replaced by worker-written ones) which were intentionally time-consuming, often involving dozens of documents.
The complete details of the question writing task are provided in Appendix~\ref{appendix:eliciting}. 


\begin{figure*}[t]\setlength{\belowcaptionskip}{-8pt}
  \centering
  \includegraphics[clip, width=0.98\textwidth]{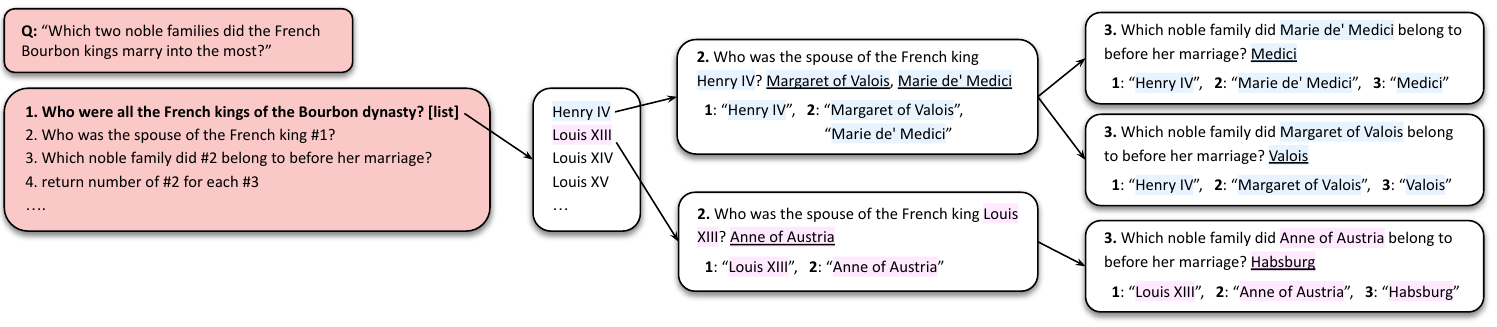}
  \vspace*{-0.4cm}
  \caption{Iteratively deriving intermediate questions using the question decomposition. The intermediate answers from previous steps (highlighted) are assigned to the decomposition step, creating the follow-up questions. The full answer assignments deriving each question are also presented at the bottom of each question box.}
  \label{fig:subquestion_derivation}
\end{figure*}

\subsection{Annotating Intermediate Answers}
\label{sec:annotation_answers}
Our next stage is to annotate each question with its corresponding answers and supporting evidence. 
Given that answering \dataset{} questions involves multiple steps and may require dozens of evidence documents, it is impractical for a single annotator to answer them directly.
To facilitate answer annotation, we first have skilled workers annotate each question with its \emph{question decomposition}, using QDMR \cite{wolfson-etal-2020-break} a widely used formalism for representing complex, multi-step questions \cite{saparina-osokin-2021-sparqling,geva-etal-2022-break,wolfson-etal-2022-weakly}. 

The decomposition is a series of intermediate steps that form a plan for answering the original question.\footnote{The same question may have multiple valid decompositions with varying levels of granularity, depending on the underlying information source. As we focus on Wikipedia, we made sure decompositions were annotated such that the answer to each intermediate question lies in a single page.} 
Part 3 of Figure~\ref{fig:monaco_overview} displays the decomposition of the question \emph{``Do people in Midwestern US states support
same sex marriage more than in the Deep South?''}. 
By following the intermediate steps, workers are only asked to answer a series of simpler questions --- while the derivation of follow-up questions and aggregation of intermediate answers is executed automatically (\S\ref{sec:annotation_execution}). As the answers to distinct intermediate questions are often found in separate sources, this also facilitates the annotation of questions whose evidence spans across many documents. For example, in Figure~\ref{fig:monaco_overview} the answers to steps 1 (\emph{``Midwestern states''}) and 4 (\emph{``states in the Deep South''}) lie in two separate documents, a table and a sentence respectively.

\begin{figure*}[t]\setlength{\belowcaptionskip}{-8pt}
  \centering
  \includegraphics[clip, width=0.98\textwidth]{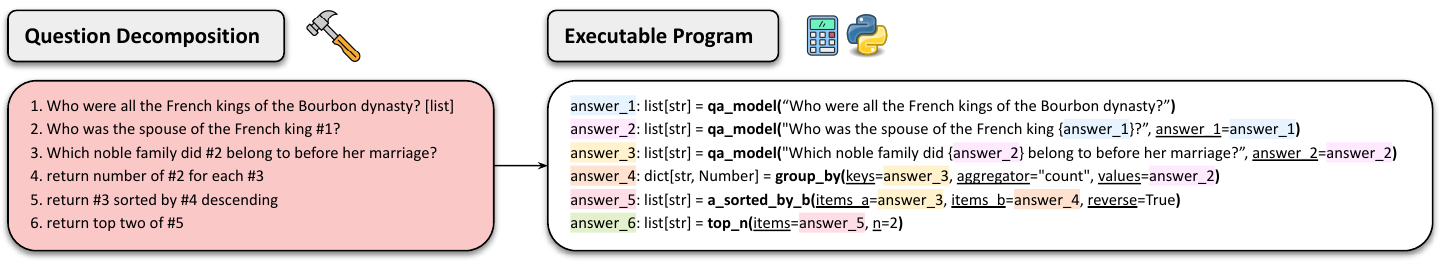}
  \vspace*{-0.4cm}
  \caption{Executing a question decomposition as a Python program. Steps 1--3 have their answers annotated by crowd workers. Steps 4--6 represent query operators and their answers are automatically computed.}
  \label{fig:decomposition_execution}
\end{figure*}

\subsection{Deriving the Final Answers}
\label{sec:annotation_execution}

To query intermediate answers and derive the final answer, we implemented a \emph{decomposition execution engine}. The role of the execution engine is twofold, as it leverages the inherent structure of the decomposition to map steps into either: (1) intermediate questions for annotation, or (2) executable Python programs that aggregate intermediate answers to produce the final answer.

First, the automatic derivation of intermediate questions, using the answers of previous steps, is presented in Figure~\ref{fig:subquestion_derivation}. The execution engine assigns the annotated answers of step \#1 (\textit{Henry IV, Louis XIII, ...}) to step \#2, thereby deriving its follow-up questions. Similarly, when deriving questions from step \#3, referring to steps 1--2, it is assigned the answers from both these steps.

The second role of the execution engine is to query the answers of intermediate steps. Namely, decomposition steps can represent different query operators that filter or aggregate the answers of previous steps. Our execution engine provides a concrete implementation for 31 such operators (Appendix~\ref{sec:appendix_executor}). It maps each operator step into an executable Python program with its input being the answers of the previous steps that it refers to.

Figure~\ref{fig:decomposition_execution} displays the decomposition of our running example and its corresponding program operators. The answers to \texttt{qa\_model} steps are annotated by workers, while the executor computes the answers of steps 4--6 based on their operator types and input (e.g. the group operator in step \#4 is given the answers of steps 2--3 as its input).

%% file: 04_dataset.tex
\section{Data Collection}
\label{sec:dataset}
This section describes the crowdsourcing process used to construct \dataset{}. As discussed in \S\ref{sec:novel_approach}, we divided data collection into several parts: question writing and decomposition (\S\ref{sec:dataset_questions}), followed by the annotation of intermediate answers and evidence (\S\ref{sec:dataset_answers_evidence}). In \S\ref{sec:dataset_validation} we describe our validation protocol to ensure the quality of our data. Lastly, estimating the potential human ``ceiling performance'' on \dataset{} is discussed in \S\ref{sec:appendix_ceiling_performance}.

\subsection{Question Writing and Decomposition}
\label{sec:dataset_questions}



As described in \S\ref{sec:question_writing}, we ``prompt'' workers to write challenging information-seeking questions that would interest different target personas. 
We employed 24 Amazon Mechanical Turk crowd-workers that were paid \$0.9 per question. Overall, we collected 1,851 questions which, following decomposition, answer annotation and validation resulted in \datasize{} fully annotated examples.

The second step in our annotation pipeline was question decomposition (\S\ref{sec:annotation_answers}). Given a question, workers were asked to write a series of intermediate steps using the QDMR syntax,
(i.e., the step templates in Appendix~\ref{sec:appendix_executor}). For non-operator steps, workers were instructed to label whether the intermediate question is expected to have a single answer or a list of answers. This label was later used to set the price of intermediate answer annotation (\S\ref{sec:dataset_answers_evidence}).
We employed four expert crowd-workers, that were already familiar with QDMR syntax, and paid them \$0.5 per decomposition.

\subsection{Intermediate Answers and Evidence}
\label{sec:dataset_answers_evidence}
Using the annotated decompositions and our execution engine, we were able to automatically derive intermediate questions which were then issued to workers (\S\ref{sec:annotation_answers}). Workers were asked to provide the answer and relevant evidence, using the English version of Wikipedia. The original (complex) question was also provided as context to help resolve any potential ambiguity. In addition, workers were asked to provide supporting evidence in the form of: (1)~the relevant Wikipedia page, (2)~the section containing the answer, and (3)~the evidence sentence, or column if the evidence is a table. If a worker determined a question is unanswerable given the information in Wikipedia, they were asked to write a short justification.
Our method ensured that all answers in \dataset{} are fully attributed to Wikipedia \cite{Bohnet2022AttributedQA} with evidence at varying levels of granularity (page, section, sentence).

Workers were paid \$0.5 for answering list questions and \$0.2 for factoid questions (single answer). We collected \datasubqsize{} intermediate questions, answers and evidence, using 10 crowd workers.

\subsection{Data Validation}
\label{sec:dataset_validation}
We placed a strong emphasis on ensuring the quality and validity of \dataset{} data. To achieve this, we employed a range of validation methods, some general, such as worker qualification and feedback mechanisms, and others tailored to the specific requirements of each annotation task.
For all of our tasks we enlisted Mechanical Turk workers from English speaking countries, with over 5,000 HITs and an approval rating higher than 95\%.
For each of the annotation tasks we held preliminary qualifications and selected the top performing workers.

\paragraph{Manual Review and Feedback}

Similar to prior benchmarks, we conducted periodic reviews for quality control \cite{yu-etal-2018-spider, zhu2024fanoutqamultihopmultidocumentquestion} and provided ongoing feedback to workers --- a practice that has been empirically shown to improve annotation quality \cite{scholman-etal-2022-design}.
All questions and decompositions (\S\ref{sec:dataset_questions}) were reviewed by one of the paper authors. 
We ensured that questions were grammatically correct, sufficiently complex and likely to be answerable using information from Wikipedia. In the question decomposition task 14\% of the annotations were corrected by the authors, primarily due to: (a) QDMR syntax errors and (b) decompositions containing questions that were unanswerable based on Wikipedia.
For intermediate answer annotation 
(\S\ref{sec:dataset_answers_evidence}) we provided ongoing feedback to maintain annotation quality, reviewing 2,079 intermediate QA tasks in total. Lastly, the authors manually validated the final answers to all of the complex questions, resulting in \datasize{} validated examples. 

\paragraph{Factoid QA Validation}
To reduce annotation costs, and motivated by the near human performance of LLMs on factoid QA tasks \cite{openai2024gpt4technicalreport, wang2024evaluating}, we chose to validate single answer questions using a hybrid approach: combining workers and a frontier LLM.

To validate worker answers we: (1) prompt an LLM for factoid QA, (2) prompt a second LLM to judge the worker-LLM answer agreement.
Using an LLM as a judge has become a standard approach \cite{phan2025humanitysexam, wei2025browsecompsimplechallengingbenchmark} as lexical matching is often insufficient for measuring answer equivalence \cite{kamalloo-etal-2023-evaluating}. 
If the LLM judged the two answers to be distinct, we re-issued the question to a second worker for annotation. The process was repeated until two answers were judged to be in-agreement. We used \gptfouro{} for both factoid QA and for judging answer agreement (prompts in Figures~\ref{fig:prompt_factoid_qa}-\ref{fig:prompt_answer_agreement}, Appendix~\S\ref{sec:appendix_prompts}). 


\paragraph{List QA Validation}
List questions are responsible for deriving multiple intermediate questions (\S\ref{sec:annotation_answers}) and unlike factoid questions, answering them remains challenging for LLMs \cite{amouyal-etal-2023-qampari, malaviya-etal-2023-quest}. We therefore validate list questions manually, having two crowd-workers annotate each question. Worker answers were then compared based on their: (1) answer overlap and (2) length difference.
The \emph{answer overlap} between lists $l_i, l_j$ is $\max\{contained(i,j), contained(j,i)\}$, where $contained(i,j) \coloneq  {|l_i \cap l_j|} / {\max\{|l_i|, |l_j|\}}$.
We consider item $k \in l_i$ to be in $l_i \cap l_j$ if there exists a $k' \in l_j$ such that the tokens of $k$ are all in the set of tokens of $k'$.
The \emph{answer length difference} of lists $l_i, l_j$ is normalized by $\max\{|l_i|, |l_j|\}$.

Answers with an overlap score higher than 0.77 and normalized length difference below 0.25 were considered to be correct, with the overlapping answers being treated as the gold answer list. If these criteria were not met, the question was re-issued to another worker for annotation, until two of the annotated lists would pass both criteria.

\paragraph{Unanswerable Questions}
Workers that were unable to answer their question, given the information in Wikipedia, labeled the question unanswerable along with a short justification. Overall, 5,363 intermediate questions were labeled as unanswerable. Of these, 2,398 questions were re-issued to a second worker, since their justifications were lacking or missing. Following re-annotation, the number of unanswerable intermediate questions was reduced to 4,083 (4.5\% of the data).

%% file: 05_data_analysis.tex
\section{The \dataset{} Dataset}
\label{sec:data_analysis}
This section provides an in-depth analysis of the \dataset{} data. In \S\ref{sec:data_analysis_stats} we break down questions based on their type, decomposition, intermediate documents and answers --- statistics that highlight the diverse nature of our data. Next, we empirically assess whether we have achieved our goal of collecting questions that are \emph{more complex} than existing benchmarks (\S\ref{sec:data_analysis_complexity}) while also \emph{more natural} than other complex QA benchmarks (\S\ref{sec:data_analysis_naturalness}).

\begin{table}[t]
\centering
\scriptsize
  \begin{tabular}{p{0.4\linewidth}r}
    \toprule
    \bf Complex questions: & \datasize{}\\ 
    \quad - Avg. \# question tokens & 14.5\\
    \quad - Avg. \# decomp. tokens & 37.6\\
    \quad - Avg. \# reasoning steps & 5.1\\
    \quad - Aggregate operators \% & 45.3\%\\
    \midrule
    \bf Intermediate questions: & 90,773\\
    \quad - List | Boolean | other & \datalistqsize{} | 40,125 | 42,099\\
    \quad - Evidence Wiki pages & 36,194\\
    \qquad - Sent. | Table | List \% & 29.5\% | 67.8\% | 2.7\%\\
    \quad - \# Wiki pages per Q & 43.3 (avg) | 12 (median)\\
    \quad - \# int. questions per Q & 66.5 (avg) | 23 (median)\\
    \quad - \# int. answers per Q & 152.5 (avg) | 53 (median)\\
  \bottomrule
\end{tabular}
\caption{\dataset{} data statistics.}
  \label{tab:monaco_stats}
\end{table}

\subsection{Data Statistics}
\label{sec:data_analysis_stats}
Table~\ref{tab:monaco_stats} summarizes the key statistics of our benchmark. \dataset{} questions are shorter than those in popular complex QA benchmarks, with 14.5 words on average compared to 17.5 in \hotpot{}, 15.7 in \musique{} and 17.3 in \fanout{}. While questions in \dataset{} are shorter, they entail far more intermediate steps than all these benchmarks (\S\ref{sec:data_analysis_complexity}). This highlights how more natural questions can often entail complex reasoning using relatively concise language.

The \datasize{} complex questions in \dataset{} have \datasubqsize{} intermediate questions, including \datalistqsize{} {\em list questions}, each list having 16.2 answers on average and a median of 5 answers. For comparison, the list QA benchmarks \qampari{} and \quest{} have 2,000 and 3,357 manually written questions, making \dataset{} the largest benchmark of human-written list questions. \dataset{} also contains 40,125 Boolean (yes/no) questions, much more than past benchmarks such as \textsc{BoolQ} \cite{clark-etal-2019-boolq} and \textsc{StrategyQA} \cite{geva-etal-2021-aristotle} with 15,942 and 2,835 respectively.

The intermediate answers are supported by evidence from 36,194 distinct Wikipedia pages. As described in \S\ref{sec:dataset_answers_evidence}, the evidence is either a sentence (29.5\%), a table (67.8\%) or a list (2.7\%). This underscores the multi-modal aspect of \dataset{}, as answering its questions requires reasoning on both paragraphs and tables \cite{talmor2021multimodalqa}.


\begingroup

\setlength{\tabcolsep}{2pt} 
\renewcommand{\arraystretch}{1} 

\begin{table}[t]
\centering
\scriptsize
  \begin{tabular}{ccccccc}
    \toprule
    Benchmark & \#Pages & \#Steps & \%Agg. & \%Arith. & \#Temp. & Divers. \\
    \midrule
    \drop & 1.0 & 2.8 & 12.5 & \bf35.0 & 39 & 2.89 \\
    \fanout & 7.0 & 3.2 & 6.0 & 2.0 & 40 & 2.59 \\
    \hotpot & 2.0 & 2.2 & 2.5 & 1.0  & 36 & 2.39 \\
    \mintaka & 1.8 & 2.4 & 12.0 & 2.0 & 33 & 2.71 \\
    \musique & 2.6 & 2.6 & 2.5 & 2.5  & 30 & 1.93  \\
    \textsc{Nat. Qs} & 1.0 & 1.1 & 2.5 & 1.0 & 13  & 0.56 \\
    \qampari  & 13.0 & 1.6 & 0.0 & 0.0  & 11 & 1.33 \\
    \quest  & 10.5 & 2.9 & 0.5 & 0.0 & 22 & 2.25 \\
    \midrule
    \dataset  & \bf43.3 & \bf 5.0 & \bf 38.0 & 17.5  & \bf 119 & \bf 4.48\\
  \bottomrule
\end{tabular}
\caption{ 
  Question complexity across benchmarks.}
  \label{tab:qa_benchmark_complexity}
\end{table}

\endgroup

\subsection{How Complex are \dataset{} Questions?}
\label{sec:data_analysis_complexity}
To measure the complexity of \dataset{} questions, we compare it to 8 popular QA benchmarks: \hotpot{} \cite{yang-etal-2018-hotpotqa}, \drop{} \cite{dua-etal-2019-drop}, \naturlquestions{} \cite{kwiatkowski-etal-2019-natural}, \mintaka{} \cite{sen-etal-2022-mintaka}, \musique{} \cite{trivedi-etal-2022-musique}, \quest{} \cite{malaviya-etal-2023-quest} and \qampari{} \cite{amouyal-etal-2023-qampari}.
Our analysis is presented in Table~\ref{tab:qa_benchmark_complexity}. 

We first compared the average number of Wikipedia pages per question (\#Pages), with \dataset{} averaging 43.3 evidence pages---more than thrice of that of the second highest benchmark.
Next, we analyzed properties pertaining to the reasoning skills that these questions entail. To assess reasoning-related properties we represented questions using their question decomposition. For \dataset{}, questions were already annotated with decompositions, while for the remaining benchmarks we generated decompositions using a few-shot prompted \gptfouro{}.
We sampled 200 questions from each benchmark's test set, generated their decompositions and measured the following:

\begin{itemize}[topsep=0pt, itemsep=0.5pt, leftmargin=.2in, parsep=0pt]
    \item \underline{Reasoning steps} (\#Steps): We list the average number of decomposition steps for the questions in each benchmark.
    \item \underline{Aggregation skills} (\%Agg.): The percentage of questions that require an aggregation operation (min, max, sum, average, median, group).
    \item \underline{Arithmetic skills} (\%Arith.): The percentage of questions entailing an arithmetic operation (addition, subtraction, division, multiplication).
    \item \underline{Reasoning templates} (\#Temp.): We define the sequence of decomposition step operators as the question's reasoning template, e.g. the template of the decomposition in Figure~\ref{fig:decomposition_execution} is:  \textit{qa\_model}*3;\textit{group\_by};\textit{a\_sorted\_by\_b};\textit{top\_n}. A higher number of reasoning templates indicates that the benchmark is more challenging.
    \item \underline{Reasoning diversity} (Divers.): We measure the diversity of questions in terms of their reasoning templates, computing Shannon's diversity index \cite{shannon1948mathematical} over the unique templates from each benchmark. 
\end{itemize}

The results in Table~\ref{tab:qa_benchmark_complexity} demonstrate that \dataset{} questions are more reasoning-intensive than past benchmarks. The average number of reasoning steps is 5.1 compared to 3.2 for the second highest benchmark. Over 38\% and 17\% of \dataset{} questions require aggregate and arithmetic operations respectively, while its diversity is significantly higher than the other benchmarks.

\subsection{How Natural are \dataset{} Questions?}
\label{sec:data_analysis_naturalness}
A main goal in building \dataset{} was to ensure that its questions were \emph{natural} and reflected the information-seeking goals of real-world users \cite{devries2020ecologicallyvalidresearchlanguage,bowman-dahl-2021-will}. We followed \citet{sen-etal-2022-mintaka} in trying to measure how natural are  \dataset{} questions compared to existing benchmarks: four containing natural questions (\naturlquestions{}, \drop{}, \mintaka{} and \fanout{}) and four with questions that were partially machine generated (\hotpot{}, \musique{}, \qampari{} and \quest{}).

Rather than having crowd-workers assign questions an underspecified (and rather ambiguous) ``naturalness'' score, we defined two criteria:
\begin{itemize}[topsep=0pt, itemsep=0.5pt, leftmargin=.2in, parsep=0pt]
    \item Labeling \textit{``who is more likely to have written the question?''} out of: (a) an expert user, (b) a regular user, (c) a machine rule-based script.
    \item A score between 1-5 indicating \textit{``how likely does the person asking the question already know its answer?''}, 5 being extremely likely.
\end{itemize}

The first criterion is fairly straightforward, while the motivation behind the second is that, in real-world settings, people posing \emph{information-seeking} questions are unlikely to be aware of the correct answer in advance. However, many questions in existing benchmarks give the impression that the question writer is already aware of the answer: \emph{``What was triggered by a British Conservative Party politician?''};\emph{``Who is the sibling of the producer of Embedded in Baghdad?''} \cite{yang-etal-2018-hotpotqa, trivedi-etal-2022-musique}. 

To score questions, we conducted a user study with 18 graduate students from the fields of natural language processing and data management. None of the paper authors took part in the user study. We randomly sampled 900 questions, with 100 questions from \dataset{} and each of the 8 aforementioned benchmarks test sets'. Each participant was asked to score 50 random questions according to the guidelines in Appendix~\ref{sec:appendix_natural_questions}. While participants were presented questions from different benchmarks they were unaware of their origin.

\begin{figure}[t]\setlength{\belowcaptionskip}{-8pt}
  \centering
  \includegraphics[clip, width=0.45\textwidth]{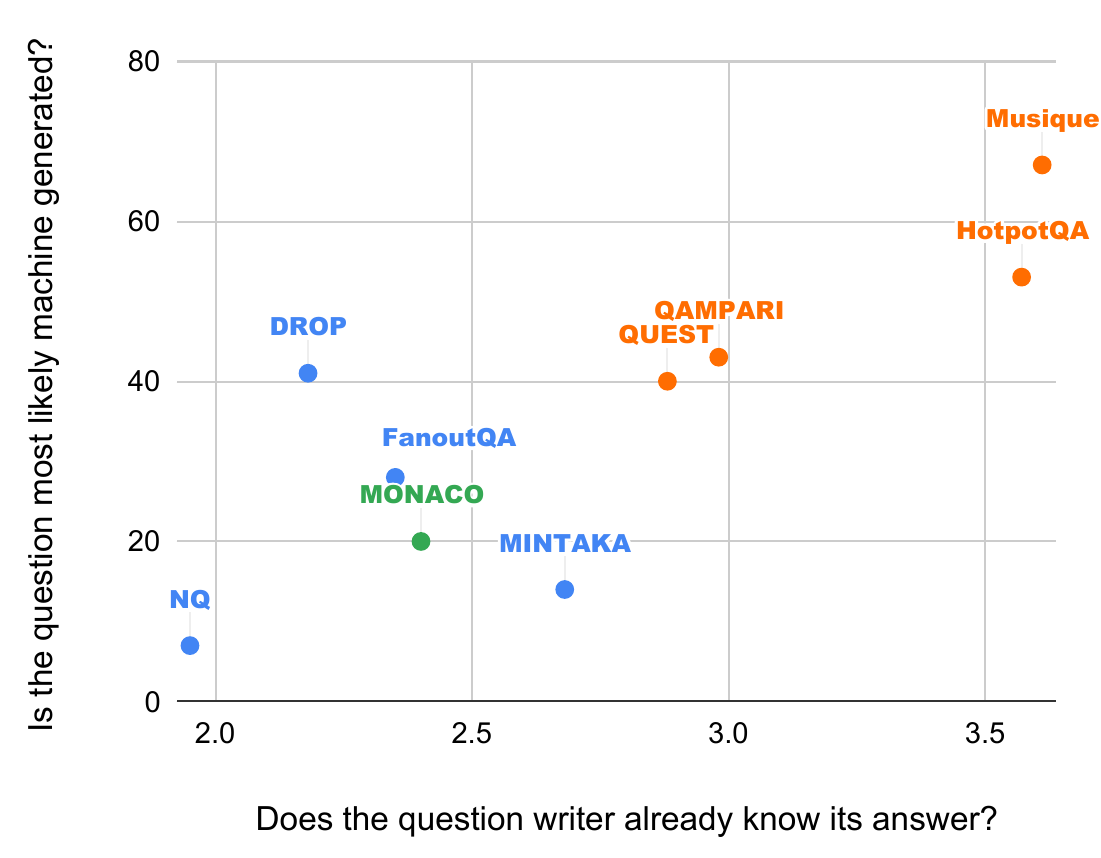}
  \vspace*{-0.4cm}
  \caption{Measuring how natural are questions from different QA benchmarks. Benchmarks with natural questions are in blue while artificial ones are in orange.} 
  \label{fig:user_study_nautralness}
\end{figure}

\begin{figure}[t]\setlength{\belowcaptionskip}{-8pt}
  \centering
  \includegraphics[clip, width=0.45\textwidth]{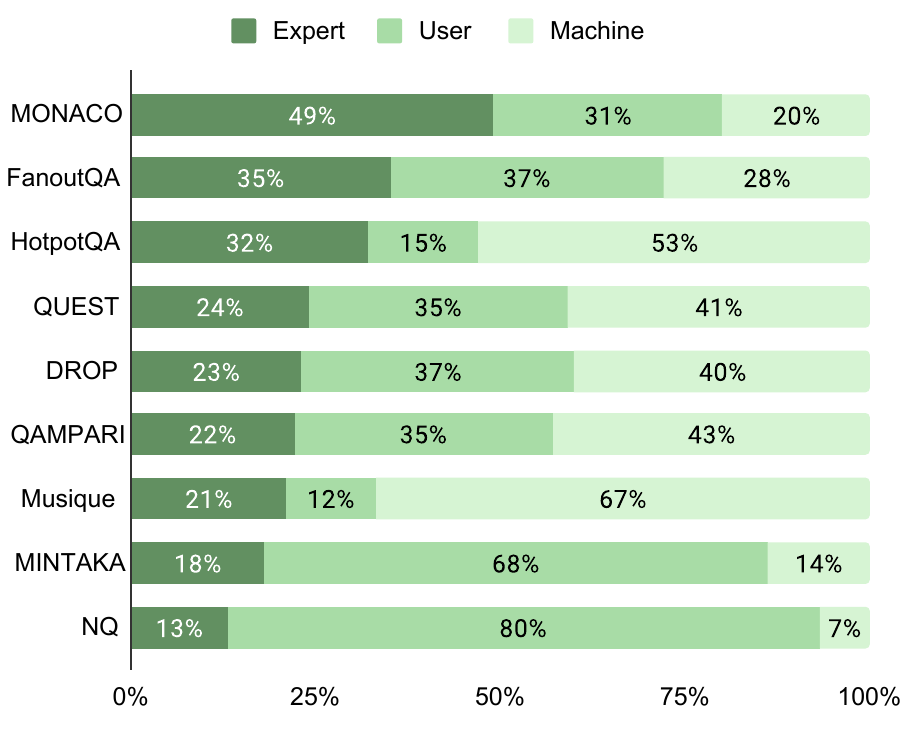}
  \vspace*{-0.4cm}
  \caption{Breaking down benchmark questions based on their most likely writer: expert, user or machine.}
  \label{fig:user_study_results}
\end{figure}

Figures \ref{fig:monaco_natural_complexity}, \ref{fig:user_study_nautralness}-\ref{fig:user_study_results} present the user study results. The ``naturalness'' score in Figure~\ref{fig:monaco_natural_complexity} is a weighted average of both our measures, normalized between 0-100. In Figure~\ref{fig:user_study_nautralness} we plot each benchmark based on the percentage of questions labeled as machine-written (y-axis) and how likely was the question writer already aware of the answer (x-axis). The results demonstrate how questions from the natural benchmarks (in blue) tend to score lower on both our measures, in stark contrast to the more artificial benchmarks (in orange). \dataset{} questions (in green) appear much closer to the natural QA benchmarks than to the artificial ones. In 
Figure~\ref{fig:user_study_results} we display the results of the most likely question writer. Somewhat unsurprising is that the top scoring Natural Questions (NQ) benchmark has only 7\% of its questions labeled as being machine-written compared to 20\% in \dataset{}. However, \dataset{} has 49\% of its questions labeled as being expert-written compared to 13\% in NQ. 

Overall, this study highlights \dataset{} as more aligned with natural QA benchmarks while being more likely to have been expert-written---results that are fully in-line with our goals.

%% file: 06_llm_models.tex
\section{Experiments}
\label{sec:models_experiments}
We evaluated 15 frontier LLMs on \dataset{} and analyzed their performance across various types of questions and prompting methods. First, we describe our evaluation setup (\S\ref{sec:models_experiments_setup}) and present LLMs' closed-book results (\S\ref{sec:models_llms_results}). In \S\ref{sec:models_experiments_breakdown} we analyze model performance based on the amount of intermediate steps and documents required to solve each question. Last, we examine LLM performance in a retrieval-augmented setting (\S\ref{sec:models_experiments_retrieval}).

\subsection{Evaluation Setup}
\label{sec:models_experiments_setup}
We evaluated 15 LLMs, including 8 models designed specifically for multi-step reasoning \cite{openai2024openaio1card, OpenAIGPTFive,deepseekai2025deepseekr1incentivizingreasoningcapability, Comanici2025Gemini2P,AnthropicClaudeFour}. All models were prompted using the same instructions, provided in Figure~\ref{fig:prompt_comlex_qa_no_cot} (Appendix~\ref{sec:appendix_prompts}). 

As string-based measures are often brittle \cite{kamalloo-etal-2023-evaluating} we use \gptfourone{}\footnote{\texttt{gpt-4.1-2025-04-14}} as a judge to validate models predictions against the gold answers. We modified the evaluation prompt used by \citet{phan2025humanitysexam, wei2025browsecompsimplechallengingbenchmark} to enable us to compute precision and recall automatically and in the case of numerical answers, to compute a normalized similarity score  (Figure~\ref{fig:prompt_llm_judge_eval}). For questions where the answer is a list of items, the LLM judge generates the total number of answers predicted by the model, followed by the list of predicted answers that also appear in the gold answers list. Both results enable us to compute precision and recall automatically based on the total number of predicted answers and the number of correctly predicted ones. 

We note that the predictions of all of the evaluated models, as well as the judgment scores generated by the LLM judge, are publicly released along with the \dataset{} data. For \claudefour{} we enabled its ``extended thinking'' with a ``budget tokens'' value of 6,000. For \geminipro{} and \geminiflash{} we asked to ``include thoughts'' using the default ``dynamic thinking'', where the model decides how much to think. \gptfive{} was evaluated using the default parameters for its ``reasoning effort'' (medium).

\subsection{Language Models Performance}
\label{sec:models_llms_results}
In this setting we evaluated the parametric knowledge of LLMs together with their ability to reason and aggregate over hundreds of facts.  
The results in Table~\ref{tab:results_llm_judge} show that all LLMs fall significantly short of achieving a perfect score on \dataset{}. Even the top performing model, OpenAI's \othree{}, reaches an F1 score of 61.2\%, leaving substantial headroom for improvement. We observe that reasoning LLMs generally outperform strong non-reasoning models such as \gptfouro{} and \llamathreeone{}. Notably, both \rone{} and \vthree{} models performed quite well and are among the strongest models evaluated.

\begin{table}[t]
\centering
\scriptsize
  \begin{tabular}{lccc}
    \toprule
    Model & Precision & Recall & F1 \\
  \midrule
    \gptfive{} (2025-08) & 66.38 & 58.98 & 60.11 \\
    \othree{} (2025-05) & \bf \underline{68.10} & \bf \underline{59.54} & \bf \underline{61.18} \\
    \ofourmini{} (2025-04) & 62.50 & 53.01 & 54.92 \\
    \othreemini{} (2025-04) & 59.29 & 46.19 & 48.75 \\
    \claudefour{} (2025-07) & 62.28 & 53.47 & 55.03 \\

    \geminipro{} (2025-07) & 65.02 & 58.14 & 59.11 \\
    \geminiflash{} (2025-07) & 58.10 & 50.60 & 52.01 \\
    \cdashline{1-4} \rule{0pt}{1em}
    \rone{} & \underline{62.52} & \underline{51.50} & \underline{53.82} \\
    \midrule
    \gptfouro{} (2025-03) & 57.37 & 46.98 & 48.98 \\
    \ \ \ + few shot chains & \underline{63.33} & \underline{52.88} & \underline{55.05} \\
    \gptfourturbo{} (2024-05) & 49.61 & 40.95 & 42.57 \\
    \ \ \ + few shot chains & 56.26 & 46.81 & 48.58 \\
    \cdashline{1-4} \rule{0pt}{1em}
    \vthree{} & 57.45 & 47.55 & 49.47 \\
    \ \ \ + few shot chains & \underline{62.31} & \underline{53.37} & \underline{55.04} \\
    \llamathreeone{} & 55.03 & 46.28 & 47.67 \\
    \ \ \ + few shot chains & 55.97 & 51.20 & 51.39 \\
    \llamathree{} & 51.40 & 43.47 & 44.76 \\
    \ \ \ + few shot chains & 55.15 & 45.12 & 47.00 \\
    \qwentwofive{} & 50.49 & 41.08 & 42.85 \\
    \ \ \ + few shot chains & 53.84 & 45.48 & 47.05 \\
    \qwentwo{} & 51.03 & 40.94 & 42.64 \\
    \ \ \ + few shot chains & 50.80 & 42.89 & 43.92 \\
  \bottomrule
\end{tabular}
\caption{Performance on \dataset{} of reasoning (top) and non-reasoning LLMs (bottom). Dashed lines separate closed-weights models from open ones. We underline the top performing model of each category.}
  \label{tab:results_llm_judge}
\end{table}

\paragraph{Chain-of-Thought}
\label{sec:models_experiments_prompting}

Prompting using a chain-of-thought (CoT) has been shown to improve performance on reasoning-heavy
tasks \cite{wei2022chain, kojima2022large}. We therefore sought to test whether CoT-prompting could further improve LLMs' performance on \dataset{}. We evaluated non-reasoning LLMs\footnote{Following OpenAI and DeepSeek guidelines, we did not experiment with CoT-prompting the reasoning LLMs.} using three CoT prompt variants: (1) a zero-shot prompt with the ``think step-by-step'' instructions; (2) a few-shot prompt with questions, answers and short explanations (Figure~\ref{fig:prompt_cot}); (3) a few-shot prompt with full reasoning chains, including all of the intermediate questions, answers and reasoning steps (Figure~\ref{fig:prompt_cot_answers}). 
For both few-shot prompts (2), (3) we included the same 21 examples which cover all of the operator types in \dataset{} (Table~\ref{tab:decomposition_ops}).

The LLMs prompted using variants (1) and (2) did not perform any better than our original non-CoT prompt. 
The key improvement we observed was in using prompt (3) where the few-shot examples contained entire reasoning chains (Figure~\ref{fig:prompt_cot_answers}, Appendix~\ref{sec:appendix_prompts}), explicitly prompting the LLM to generate all intermediate reasoning steps before its answer. 
Ultimately, the results in Table~\ref{tab:results_llm_judge} show that all non-reasoning LLMs improve when prompted with few-shot reasoning chains.

\subsection{Performance Breakdown and Analysis}
\label{sec:models_experiments_breakdown}

\begin{figure*}[t]\setlength{\belowcaptionskip}{-8pt}
  \centering
  \includegraphics[clip, width=0.36\textwidth]{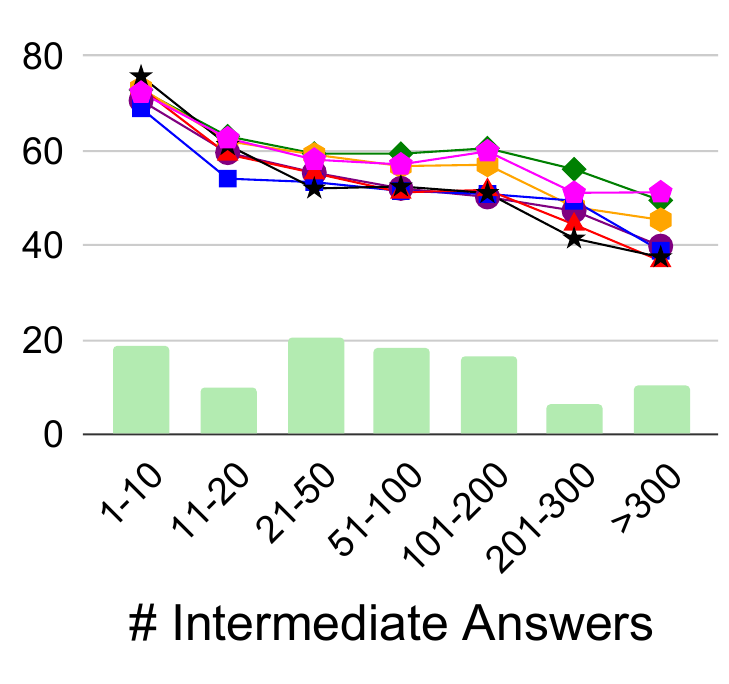}
  \vspace*{-0.4cm}
  \includegraphics[clip, width=0.52\textwidth]{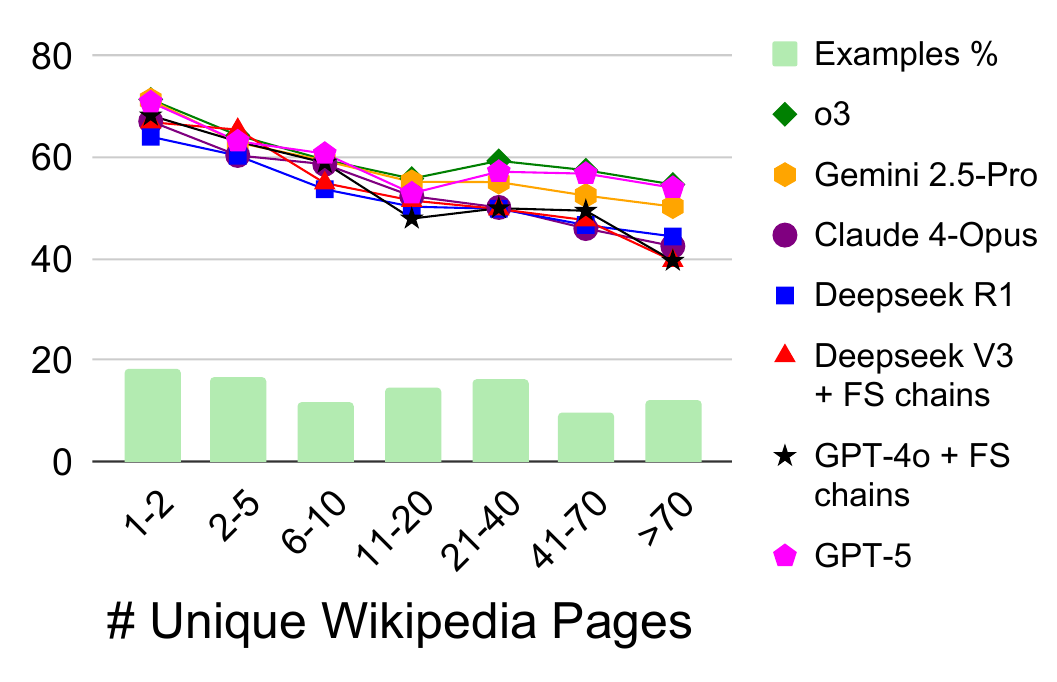}
  \caption{Complex QA performance (F1) as a function of the number intermediate answers / evidence pages.}
  \label{fig:results_complex_qa_decomp_steps}
\end{figure*}

\paragraph{Breakdown by Question Properties}
We break down model performance based on the amount of intermediate steps and factual information entailed by \dataset{} questions. Namely, we group questions based on the number of: (1) intermediate answers, and (2) distinct Wikipedia pages. Figure~\ref{fig:results_complex_qa_decomp_steps} presents the performance breakdown results, highlighting how all models struggle as the number of intermediate answers and evidence increases. However, \othree{} and \gptfive{} performance is an exception: while it is on par with the other LLMs when the number of pages or intermediate answers is small, the gap increases significantly as the number of Wiki pages is in the dozens (>20).

\begin{figure}[t]\setlength{\belowcaptionskip}{-8pt}
  \centering
  \includegraphics[clip, width=0.45\textwidth]{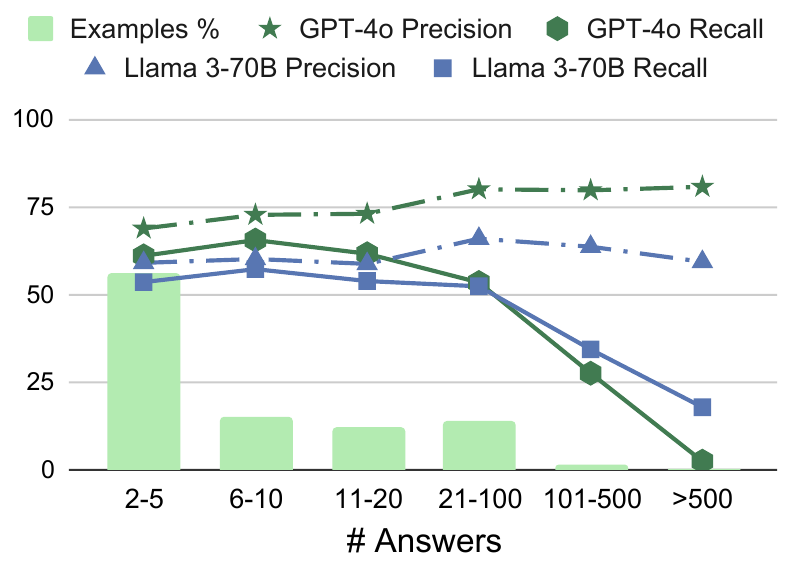}
  \vspace*{-0.4cm}
  \caption{List QA prompted LLMs performance on the intermediate list questions in \dataset{}. We provide the average precision and recall scores. The full results are provided in Appendix~\ref{sec:appendix_listqa_performance}}
  \label{fig:results_list_qa}
\end{figure}

\paragraph{List Questions} 
Questions in \dataset{} generally entail solving intermediate steps that have a list of answers. List questions have been shown to be challenging for past LLMs \cite{amouyal-etal-2023-qampari, malaviya-etal-2023-quest} and we revisit this challenge by evaluating \gptfouro{} and \llamathree{} on all \datalistqsize{} intermediate list question in \dataset{}. Both models were prompted to generate an exhaustive list of answers using the few-shot prompt in Figure~\ref{fig:prompt_list_qa} (Appendix~\ref{sec:appendix_prompts}). Due to cost limitations, we evaluate list QA precision and recall using string-based measures: matching each predicted answer to its most similar item in the gold answers based on their tokens F1 score. The results in Figure~\ref{fig:results_list_qa} demonstrate that list QA remains challenging for frontier LLMs. While \gptfouro{} precision is high (75-80\%) its recall sharply decreases with the number of expected answers. 

\subsection{Retrieval Augmented LLMs}
\label{sec:models_experiments_retrieval}
\begin{figure}[t]\setlength{\belowcaptionskip}{-8pt}
  \centering
  \includegraphics[clip, width=0.45\textwidth]{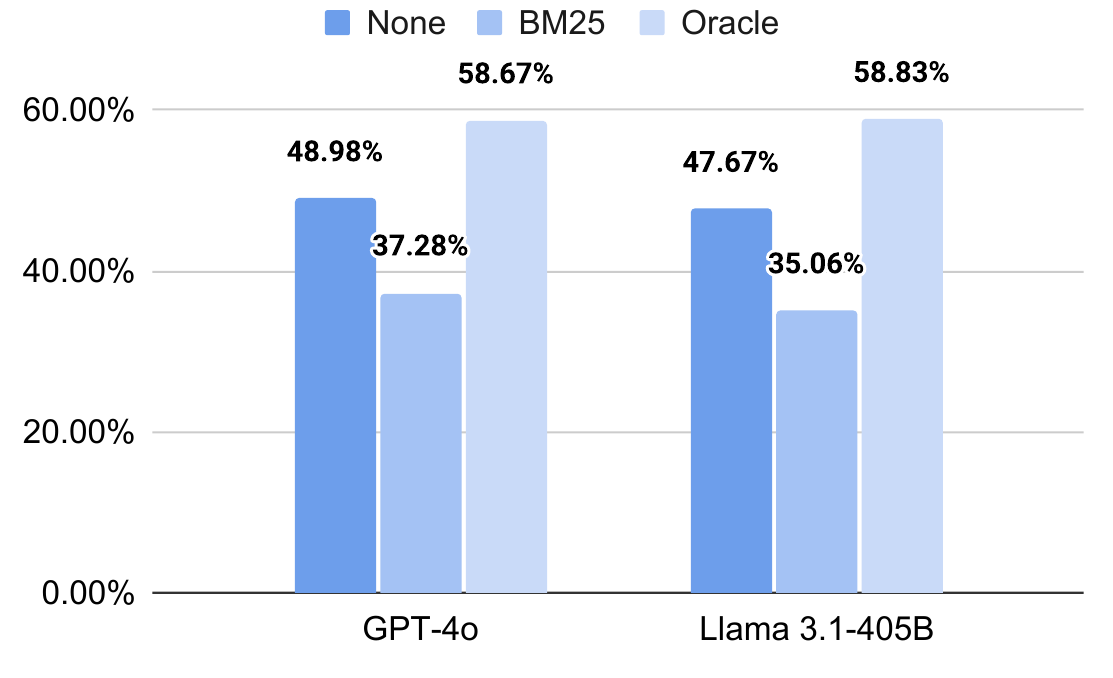}
  \vspace*{-0.48cm}
  \caption{Retrieval-augmented LLM performance on \dataset{} (F1 score), with documents retrieved by BM25 or given the gold evidence documents (Oracle). The full results are provided in Appendix~\ref{sec:appendix_retrieval_performance}}
\label{fig:results_retrieval_complex_qa}
\end{figure}

\paragraph{Oracle Retrieval}
Unlike the experiments in \S\ref{sec:models_llms_results}, where models rely solely on their parametric knowledge, we also evaluated the complex reasoning skills of models over long contexts, when all the relevant evidence to \dataset{} questions is provided as input. 
This is an \emph{Oracle retrieval} setting, where the LLM is provided all of the gold evidence documents in its prompt, effectively evaluating its reasoning skills in isolation from the task of knowledge retrieval. We evaluate \gptfouro{} and \llamathreeone{}, and use the prompt in Figure~\ref{fig:prompt_retrieval_augmented} (Appendix~\ref{sec:appendix_prompts}) to indicate that all input documents are indeed relevant. The Oracle results in Figure~\ref{fig:results_retrieval_complex_qa} show that both LLMs experience a 10 point improvement compared to the closed-book setting (None). Nevertheless, the fact that they reach only 58.7\% indicates that even with all relevant knowledge, the complex reasoning in \dataset{} questions remains a challenge.

\paragraph{End-to-end RAG}

Retrieval-augmented generation \cite{lewis2020retrieval} is a highly realistic setting where the LLM is provided with retrieved documents. We constructed a new retrieval index for Wikipedia, with 80M documents, since unlike the paragraph-specific indices used in prior work \cite{kwiatkowski-etal-2019-natural}, documents in \dataset{} can be either paragraphs, tables or lists. Our retriever is BM25 \cite{robertson2009probabilistic} and we provide its top-20 documents as input to the LLM prompt (Figure~\ref{fig:prompt_retrieval_augmented}, Appendix~\ref{sec:appendix_prompts}). Ideally, RAG should improve over the closed-book setting, with LLMs utilizing information outside of their parametric knowledge while ignoring any irrelevant documents. However, both RAG models experience a sharp drop in performance compared to the original closed-book setting (Figure~\ref{fig:results_retrieval_complex_qa}). While BM25 documents may only be partially relevant, our RAG prompt explicitly instructs to ignore the retrieved documents in such cases. While this lack of ``retrieval robustness'' \cite{yoran2024making} has been observed in weaker models, the limitation appears to persist even in current frontier LLMs.

%% file: 07_related_work.tex
\section{Related Work}
\label{sec:related_work}

Information-seeking question answering is a longstanding approach to evaluate the reasoning skills of LLMs \cite{rodriguez-boyd-graber-2021-evaluation}.
Natural QA benchmarks consist almost entirely of ``simple'' questions, where the answers can be found within a single span of text \cite{nguyen2016ms,Abujabal2018ComQAAC, jin-etal-2019-pubmedqa, krithara2023bioasq,wei2024measuringshortformfactualitylarge}. As the reasoning skills of models improved, researchers have sought to evaluate LLMs on questions that require multi-hop reasoning skills \cite{yang-etal-2018-hotpotqa,geva-etal-2021-aristotle,trivedi-etal-2022-musique,press-etal-2023-measuring,wei2025browsecompsimplechallengingbenchmark}, as well as questions whose answer is a set of items which is spread across many documents \cite{amouyal-etal-2023-qampari, malaviya-etal-2023-quest, zhong-etal-2023-romqa, katz-etal-2023-neretrieve, rassin-etal-2024-evaluating}. 

Recent works introduced benchmarks featuring more naturally elicited questions that involve multiple documents \cite{sen-etal-2022-mintaka, zhu2024fanoutqamultihopmultidocumentquestion, malaviya-etal-2024-expertqa,krishna2024factfetchreasonunified,yang2024crag, phan2025humanitysexam}. However, solving the questions in each of these benchmarks involves no more than a handful of webpages. By contrast, \dataset{} questions are significantly more time-consuming, as the evidence to answer each question typically involves dozens of pages.

More recently, \citet{wei2025browsecompsimplechallengingbenchmark} introduced the BrowseComp benchmark for retrieving hard-to-find information. \dataset{} differs from this work in two main aspects. First, BrowseComp questions target niche topics and are phrased more as trivia questions, where the question writer already knows the answer and is trying to challenge someone else. In contrast, \dataset{} questions aim to convey the info-seeking goals of real-world users (\S\ref{sec:data_analysis_naturalness}). Second, the answers of BrowseComp questions are easy to verify, as their supporting evidence involves only a few webpages. \dataset{} questions are much broader, with answers that require to synthesize many pieces of evidence that are spread across dozens of Wikipedia pages.

WideSearch, a contemporary work by \citet{wong2025widesearchbenchmarkingagenticbroad}, is a benchmark of real-world questions that evaluate LLM agents' ability to collect broad information. WideSearch contains 100 English questions, compared to 1,350 in \dataset{}.

Another key aspect in QA research has been benchmarking retrieval-augmented LLMs \cite{lewis2020retrieval,ram-etal-2023-context}. A great number of retrieval benchmarks focus on information-seeking tasks that are either domain-general \cite{voorhees2000building,thakur2021beir,yang2024crag}, domain-specific \cite{dasigi-etal-2021-dataset,openscholar} or involve implicit reasoning skills \cite{geva-etal-2021-aristotle, hongjin2024bright}. Nevertheless, the questions in all of these benchmarks rarely require to retrieve more than five evidence documents. The sheer breadth of evidence in \dataset{} position it as an advanced and more rigorous benchmark than many of the existing tasks for retrieval-augmented generation \cite{hsieh2024ruler, krishna2024factfetchreasonunified} and factual attribution \cite{Bohnet2022AttributedQA,gao-etal-2023-enabling, jacovi2025factsgroundingleaderboardbenchmarking}.

%% file: 08_conclusion.tex
\section{Conclusion}
\label{sec:conclusion}
Questions that require extensive research across multiple webpages are incredibly important, yet they are not represented in current LLM benchmarks.
To this end we introduce \dataset{}, a benchmark of human-written, realistic and time-consuming questions. It contains \datasize{} questions, manually annotated with gold standard reasoning chains that include \datasubqsize{} intermediate questions, answers and supporting evidence documents. The modest performance of frontier LLMs on \dataset{} suggests that reasoning over dozens of documents remains an open challenge. As such, \dataset{} serves as a unique testbed for evaluating LLMs-powered systems on much broader tasks, which span across lots of documents and demand extensive factual knowledge, information retrieval and reasoning skills.

%% file: 09_limitations.tex
\section{Limitations}
\label{sec:limitations}

\paragraph{Time-dependent Questions}
A key factor in the evaluation of information-seeking questions is the potential of certain answers to change over time \cite{zhang-choi-2021-situatedqa, vu-etal-2024-freshllms}. In our experiments, we did not measure the effect that temporal-dependence may have on the accuracy of future models evaluated on \dataset{}. Nevertheless, we fully provide users with the necessary resources to account for time-dependent answers.\footnote{\url{https://huggingface.co/datasets/allenai/MoNaCo_Benchmark}} Each question in \dataset{} was manually labeled by the paper authors as to whether its answer is expected to: (a) remain as is; (b) change every few years; (c) change on a yearly basis. Overall, 49.8\% of the questions are time-independent; 34.1\% have answers that change every few years; and 16.1\% of have answers that are expected to change each year. Furthermore, we release the full timestamp of each answer in \dataset{} (i.e. the answer annotation date), enabling users to determine the relevant time-frame for every question in our data. This data should facilitate future usage of \dataset{} and enables researchers to evaluate models' performance at particular points in time.

\paragraph{Deep Research Systems}
In our experiments, we did not evaluate any LLM-powered ``Deep Research'' systems \cite{OpenAIDeepRS,Huang2025DeepRA} nor did we evaluate multi-step RAG systems that iteratively retrieve, decompose and answer complex questions --- interleaving retrieval and QA \cite{asai2023selfrag, trivedi-etal-2023-interleaving, yoran-etal-2023-answering}. Since answering \dataset{} questions requires to retrieve and synthesize information from dozens of different webpages, it can serve as a useful testbed for evaluating the capabilities of such Deep Research systems.

\paragraph{LLM as a Judge}
As discussed in \S\ref{sec:models_experiments_setup}, we used \gptfourone{}, to compare the answers generated by models to the gold answers and to determine whether or not the two are equivalent. While this is a commonly used practice \cite{phan2025humanitysexam, wei2025browsecompsimplechallengingbenchmark} the judge model may still make mistakes, predicting wrong answers to be correct and vice-versa. Nevertheless, we view this as a reasonable compromise and as LLMs continue to improve, swapping \gptfourone{} for a ``stronger'' judge model should further improve future evaluation.


%% file: 00_appendix.tex
\appendix
\section{Appendix: Supplementary Results}
\input{appendix_ceiling_performance}
\input{appendix_list_qa}
\input{appendix_retrieval_performance}

\newpage
\section{Appendix: Additional Implementation Details}

\input{appendix_question_writing}

\input{appendix_qdmr_operators}
\input{appendix_user_study}

\input{appendix_prompt}

\fi

%% file: appendix_ceiling_performance.tex
\subsection{Benchmark Ceiling Performance}
\label{sec:appendix_ceiling_performance}

Several question answering benchmarks have included estimates of the human ceiling performance on their task \cite{rajpurkar-etal-2016-squad,yang-etal-2018-hotpotqa, dua-etal-2019-drop,geva-etal-2021-aristotle}. Such benchmarks typically contain questions which are not time-consuming and can be solved relatively fast by a single human annotator. By contrast, \dataset{} questions are time-consuming and require to combine information from 43 distinct webpages on average. This inherent complexity is what encouraged us to divide our benchmark's annotation process into multiple, much simpler steps, as described in \S\ref{sec:dataset}.

Rather than estimating human performance directly, we sought to ensure that the collected answers to \dataset{} questions were indeed valid (\S\ref{sec:dataset_validation}). Question decompositions were manually reviewed; factoid question answers were human-annotated, then validated by \gptfouro{}, or a second crowd-worker if the first worker's answer contradicted that of the LLM. In the case of intermediate ``list questions'' there was a higher risk of ambiguity in the question. For example, in questions like \textit{``What are some vegan dishes in Chile?''} or \textit{``Who are female poets of Andalusian Spain?''}. However, the answers to such questions are unambiguous given our underlying information-source --- the English version of Wikipedia. Furthermore, the validation process of list QA demanded a high answer overlap between two human annotators (>77\%), which further indicates our list questions are unambiguous, given Wikipedia.

Lastly, the execution of operator steps such as aggregation or numerical comparison is performed automatically, leaving no room for potential calculation errors by humans.

%% file: appendix_list_qa.tex
\subsection{List QA Performance}
\label{sec:appendix_listqa_performance}

Table~\ref{tab:results_list_qa} provides the detailed performance breakdown of models over the intermediate list questions in \dataset{}. The results are displayed as a plot in Figure~\ref{fig:results_list_qa}.

\begin{table}[h]
\centering
\scriptsize
  \begin{tabular}{cccccc}
    \toprule
    Model & \# Ans. & \# Ex. & Pre. & Rec. & F1 \\
    \midrule
    \multirow{7}{*}{\gptfouro{}} & 2-5 & 4,812 & 68.9 & 61.1 & 62.3 \\
  & 6-10 & 1,312 & 72.8 & 65.6 & 66.5 \\
  & 11-20 & 1,064 & 73.1 & 61.7 & 64.5  \\
  & 21-100 & 1,206 & 80.1 & 53.5 & 60.7 \\
  & 101-500 & 137 & 79.8 & 27.6 & 37.6 \\
  & >500 & 18 & 80.8 & 2.5 & 4.8 \\
  & \underline{All}: & 8,549 & 76.0 & 61.8 & 64.5 \\
  \midrule
    \multirow{7}{*}{\llamathree{}} & 1-5 & 4,812 & 59.1 & 53.9 & 53.6 \\
  & 6-10 & 1,312 & 60.2 & 59.8 & 57.3 \\
  & 11-20 & 1,064 & 58.8 & 53.8 & 53.9  \\
  & 21-100 & 1,206 & 66.0 & 47.6 & 52.4 \\
  & 101-500 & 137 & 63.7 & 28.2 & 34.4 \\
  & >500 & 18 & 59.4 & 19.0 & 17.9 \\ 
  & \underline{All}: & 8,549 & 60.3 & 53.3 & 53.7 \\

  \bottomrule
\end{tabular}
\caption{List QA prompted LLMs performance on the intermediate list questions in \dataset{}. We provide the average precision, recall and F1 scores.}
  \label{tab:results_list_qa}
\end{table}

%% file: appendix_retrieval_performance.tex
\subsection{Retrieval Performance}
\label{sec:appendix_retrieval_performance}

Table~\ref{tab:results_retrieval_complex_qa} provides the detailed results of the RAG experiments on \dataset{} (also plotted in Figure~\ref{fig:results_retrieval_complex_qa}).

\begin{table}[h]
\centering
\scriptsize
  \begin{tabular}{cccccc}
    \toprule
    Model & Retriever & Pre. & Rec. & F1 \\
    \midrule
    \multirow{3}{*}{\gptfouro{}} & None & 57.37 & 46.98 & 48.98 \\
  & BM25 & 48.21 & 34.94 & 37.28 \\
  & Oracle & 67.28 & 56.08 & 58.67  \\
  \midrule
    \multirow{3}{*}{\llamathreeone{}} & None & 55.03 & 46.28 & 47.67 \\
  & BM25 & 44.18 & 33.32 & 35.06 \\
  & Oracle & 66.68 & 56.57 & 58.83 \\

  \bottomrule
\end{tabular}
\caption{LLM performance on \dataset{} given gold evidence documents as input.}
\label{tab:results_retrieval_complex_qa}
\end{table}

%% file: appendix_question_writing.tex
\subsection{Eliciting Complex Natural Questions: Additional Details}
\label{appendix:eliciting}

In each question-writing task, workers were provided with: (1) a persona description, (2) helpful keywords, and (3) reference questions from different personas. We came up with 28 target personas, covering a wide range of domains: geography, demographics, politics, world leaders, economics, higher education, languages, history, nutrition, cuisine, sports, film, television, music, literature and theater. Each persona was \emph{described} in 2-4 sentences, as shown in Part 1 of Figure~\ref{fig:monaco_overview}. To promote diversity, workers received seven \emph{keywords} which were randomly selected from a larger persona-specific pool. To encourage the creation of challenging questions, we also included five \emph{reference questions} (initially written by us and later replaced by worker-written ones) which were intentionally time-consuming, often involving dozens of documents. Importantly, we were careful to use reference questions from personas that were \emph{unrelated} to the target persona, thereby discouraging workers from simply paraphrasing reference questions. 
To further promote creativity, we avoided restricting workers to a set of pre-defined question templates. Finally, in order to reduce any potential reasoning shortcuts in questions, we did not provide workers with any of the answers (or evidence) in advance. 

Overall, this approach enabled us to collect questions that were judged as both realistic and significantly more time-consuming than those in existing benchmarks \S\ref{sec:data_analysis_naturalness}.

%% file: appendix_qdmr_operators.tex
\subsection{Decomposition Execution}
\label{sec:appendix_executor}
Table~\ref{tab:decomposition_ops} lists all query operators supported by our executor and their corresponding Python function signatures. This implementation extends the original QDMR representation in \citet{wolfson-etal-2020-break} to 31 operators.

\begin{table*}[h]
\tiny
  \begin{tabular}{p{0.12\linewidth}p{0.27\linewidth}p{0.57\linewidth}}
    \toprule
    \bf Operator & \bf Decomposition Template & \bf Python Function Signature \\
    \midrule
    
    Question & \textit{A free-form question} & {\fontfamily{qcr}\selectfont qa\_model(question: str) -> str | list[str]} \\\midrule
    
    Yes/no filter & return \#x where \#y is \{true / false\} & {\fontfamily{qcr}\selectfont filter\_boolean(entities: list[str], booleans: list[bool], required\_value: bool = True) -> list[str]} \\\midrule
    
    Comparison filter & return \#x where \#y \{comparator\} \{val\} & {\fontfamily{qcr}\selectfont filter\_compare(entities: list[str], left\_values: list[float], comparator: Literal[">", "<", ">=", "<=", "=="],        right\_value\_or\_values: Number | datetime | list[float] | list[datetime]) -> list[str]:} \\\midrule

    Superlative filter & return \#x where \#y is \{highest / lowest\}
 & {\fontfamily{qcr}\selectfont filter\_superlative(entities: list[str], values: list[float], superlative: Literal["max", "min"]) -> list[str]} \\\midrule

    Sort & return \#x sorted \{by \#y\} \{ascending / descending\} & {\fontfamily{qcr}\selectfont a\_sorted\_by\_b(items\_a: list[str], items\_b: list[Number | datetime], reverse: bool = False) -> list[str]} \\\midrule

    Set intersection & return \{\#x / val\} in both \#y and \#z & {\fontfamily{qcr}\selectfont items\_in\_both(list\_of\_items\_a: list[str], list\_of\_items\_b: list[str]) -> list[str]} \\\midrule

    Set union & return \#x, \#y \{, \#z, ...\} & {\fontfamily{qcr}\selectfont concatenate\_items(*items: str | Number | datetime) -> list[str | Number | datetime]} \\\midrule

    Set difference & return \#x besides \{ \#y / val\} & {\fontfamily{qcr}\selectfont discard(items: list[str], items\_to\_discard: str | list[str]) -> list[str]} \\\midrule

    Top-n & return the top \{num\} of \#x & {\fontfamily{qcr}\selectfont top\_n(items: list[str], n: int) -> list[str]} \\\midrule

    Item at n'th position & return the \{ordinal\} of \#x & {\fontfamily{qcr}\selectfont access\_list\_index(items: list[str], n: int) -> str} \\\midrule

    Addition & return sum of \#x and \#y & {\fontfamily{qcr}\selectfont addition(number\_or\_list\_1: Number | list[Number], number\_or\_list\_2: Number | list[Number]) -> Number | list[Number]} \\\midrule

    Subtraction & return difference of \#x and \#y & {\fontfamily{qcr}\selectfont difference(number\_or\_list\_1: Number | list[Number], number\_or\_list\_2: Number | list[Number]) -> Number | list[Number]} \\\midrule

    Multiplication & return multiplication of \#x and \#y & {\fontfamily{qcr}\selectfont multiplication(number\_or\_list\_1: Number | list[Number], number\_or\_list\_2: Number | list[Number]) -> Number | list[Number]} \\\midrule

    Division & return division of \#x and \#y & {\fontfamily{qcr}\selectfont division(number\_or\_list\_1: Number | list[Number], number\_or\_list\_2: Number | list[Number]) -> Number | list[Number]} \\\midrule

    Percentage & return percentage of \#x and \#y & {\fontfamily{qcr}\selectfont percentage(number\_or\_list\_1: Number | list[Number], number\_or\_list\_2: Number | list[Number]) -> Number | list[Number]} \\\midrule

    Count & return number of \#x & {\fontfamily{qcr}\selectfont count(items: list[str]) -> int} \\\midrule

    Average & return average of \#x & {\fontfamily{qcr}\selectfont average(items: list[Number]) -> Number} \\\midrule

    Median & return median of \#x & {\fontfamily{qcr}\selectfont median(items: list[Number]) -> Number} \\\midrule

    Max & return highest of \#x & {\fontfamily{qcr}\selectfont max(items: list[Number]) -> Number} \\\midrule

    Min & return lowest of \#x & {\fontfamily{qcr}\selectfont min(items: list[Number]) -> Number} \\\midrule

    Sum & return sum of \#x & {\fontfamily{qcr}\selectfont sum(items: list[Number]) -> Number} \\\midrule

    Group & return  the \{aggregate\} of \#y for each \#x & {\fontfamily{qcr}\selectfont group\_by(entities: list[str], aggregator: Literal["count", "sum", "average", "median", "min", "max"], values: list[Number | datetime]) -> dict[str, Number | datetime] | list[str, Number | datetime]} \\\midrule

    == & return if \#x is equal to \{\#y / val\} & {\fontfamily{qcr}\selectfont equals(item\_1: str | Number | datetime, item\_2: str | Number | datetime) -> bool} \\\midrule

    < & return if \#x is less than  \{\#y / val\} & {\fontfamily{qcr}\selectfont greater\_than(item\_1: Number | datetime, item\_2: Number | datetime) -> bool} \\\midrule

    > & return if \#x is higher than  \{\#y / val\} & {\fontfamily{qcr}\selectfont less\_than(item\_1: Number | datetime, item\_2: Number | datetime) -> bool} \\\midrule

    >= & return if \#x is at most  \{\#y / val\} & {\fontfamily{qcr}\selectfont at\_least(item\_1: Number | datetime, item\_2: Number | datetime) -> bool} \\\midrule

    <= & return if \#x is at least  \{\#y / val\} & {\fontfamily{qcr}\selectfont at\_most(item\_1: Number | datetime, item\_2: Number | datetime) -> bool} \\\midrule

    Boolean intersection & return if both \#x and \#y are \{true / false\} & {\fontfamily{qcr}\selectfont both\_true(items\_a: bool, items\_b: bool) -> bool} \\\midrule

    Arg max & return which is highest of \#x, \#y \{, \#z, ...\} & {\fontfamily{qcr}\selectfont argmax(*items: Number) -> Number} \\\midrule

    Arg min & return which is lowest of \#x, \#y \{, \#z, ...\} & {\fontfamily{qcr}\selectfont argmin(*items: Number) -> Number} \\\midrule

    Arg true & return which is true of \#x, \#y \{, \#z, ...\} & {\fontfamily{qcr}\selectfont which\_is\_true(items\_a: bool, items\_b: bool) -> bool} \\

  \bottomrule
\end{tabular}
\caption{All of the query operators supported by our decomposition executor.}
\label{tab:decomposition_ops}
\end{table*}

%% file: appendix_user_study.tex
\subsection{Natural Questions User Study}
\label{sec:appendix_natural_questions}
\begin{figure*}[h]\setlength{\belowcaptionskip}{-8pt}
  \centering
  \includegraphics[clip, width=0.47\textwidth]{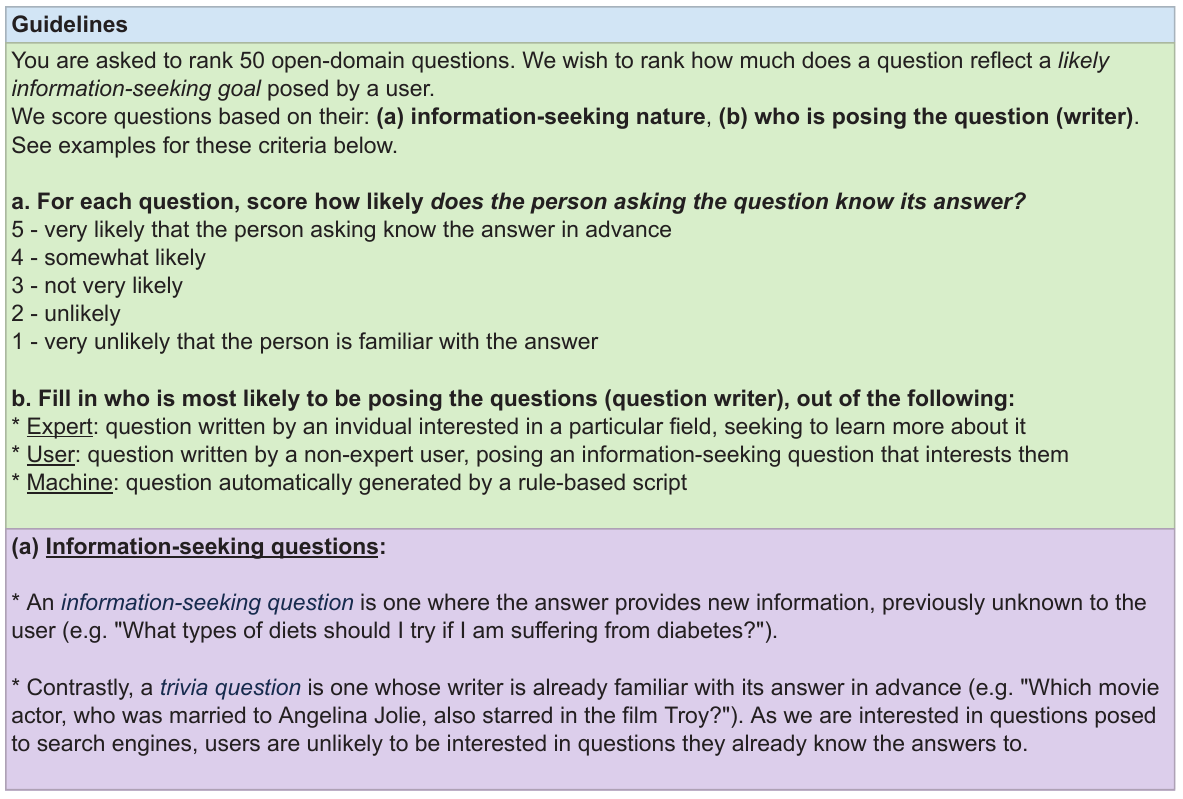}
  \vspace*{-0.1cm}
  \includegraphics[clip, width=0.47\textwidth]{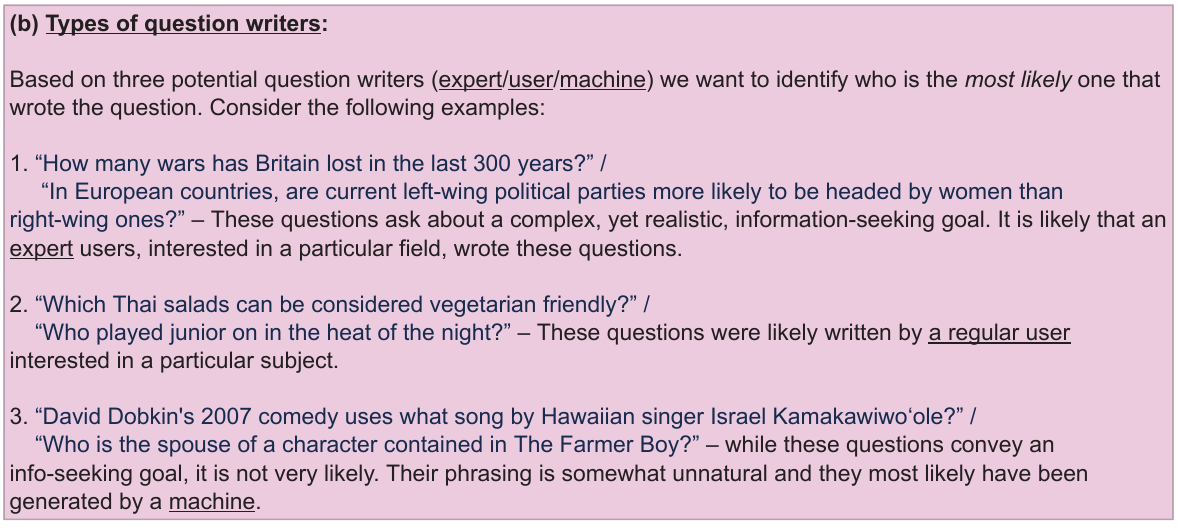}
  \vspace*{-0.4cm}
  \caption{The guidelines for our user study measuring how natural are the questions of complex QA benchmarks.}
  \label{fig:naturalness_user_study}
\end{figure*}

Figure~\ref{fig:naturalness_user_study} provides the full guidelines given to the participants of our user study that measures how natural are the questions in \dataset{}, compared to those in other information-seeking QA benchmarks. For the user study details see \S\ref{sec:data_analysis_naturalness}.

%% file: appendix_prompt.tex
\subsection{Prompts}
\label{sec:appendix_prompts}

This section describes the prompts used throughout the paper. Please note that some of the prompts have only part of their examples listed, to save space. For the full prompt descriptions please refer to our project's repository at: \url{https://tomerwolgithub.github.io/monaco/}.

\begin{figure*}[t]\setlength{\belowcaptionskip}{-8pt}
\footnotesize
  \begin{tabular}{p{0.97\linewidth}}
    \toprule \textbf{System prompt:}\\
    You are a helpful question answering assistant. Your task is to answer a complex question provided by the user. You may generate an explanation before providing the answer. The answer must be generated as a concise list of one or more entities, numbers or dates. You must always answer the question, even if your information is not up-to-date, please answer based on it. \\

Your response must use the following format: \\
Answers: \{ANSWERS\}\\
Where ANSWERS is a list of potential answers, separated by commas. You must end your response after the final answer. You must always answer the question, even if your information is not up-to-date, please answer based on it. \vspace{0.005mm}\\

\bottomrule
\end{tabular}
\caption{Complex QA system prompt, used to benchmark LLMs performance on \dataset{}.}
\label{fig:prompt_comlex_qa_no_cot}
\end{figure*}

\begin{figure*}[t]\setlength{\belowcaptionskip}{-8pt}
\footnotesize
  \begin{tabular}{p{0.97\linewidth}}
    \toprule \textbf{System prompt:}\\
    Judge whether the following [response] to [question] is correct or not based on the precise and unambiguous [correct\_answer] below. 
    
[question]: \{question\}

[response]: `\{response\}'
\\\vspace{0.005mm}

Your judgment must be in the format and criteria specified below:
\\\vspace{0.005mm}

\underline{extracted\_final\_answer}: The final exact answer extracted from the [response]. Put the extracted answer as `None’ if there is no exact final answer to extract from the response.
\\\vspace{0.005mm}

[correct\_answer]: \{correct\_answer\}
\\\vspace{0.005mm}

\underline{final answer length}: Provide the overall number of unique answers that appear in [response], not just the correct ones. Be sure to provide a number, not an estimate! 
\\\vspace{0.005mm}

\underline{reasoning}:  Explain why the extracted\_final\_answer is correct or incorrect based on [correct\_answer], focusing only on if there are meaningful differences between [correct\_answer] and the extracted\_final\_answer. Do not comment on any background to the problem, do not attempt to solve the problem, do not argue for any answer different than [correct\_answer], focus only on whether the answers match.
\\\vspace{0.005mm}

\underline{correct}: Answer `yes’ if extracted\_final\_answer matches the [correct\_answer] given above, or is within a small margin of error for numerical problems, a margin of 1 to 5.5 percentage points is acceptable. Answer `no’ otherwise, i.e. if there is any inconsistency, ambiguity, non-equivalency, or if the extracted answer is incorrect.
\\\vspace{0.005mm}

\underline{precision}: Answer `1’ if extracted\_final\_answer matches the [correct\_answer] given above. Answer `0’ otherwise, i.e. if there is any inconsistency, ambiguity, non-equivalency, or if the extracted answer is incorrect. In the case where [correct\_answer] is a number or percentage, then answer with the following formula to compute the normalized similarity score: 

[1 - (abs([correct\_answer] - extracted\_final\_answer) / max(abs([correct\_answer]), abs(extracted\_final\_answer)))]
\\\vspace{0.005mm}

\underline{final precision}: Extract the precision score from above, just the final score (number).
\\\vspace{0.005mm}

\underline{overlapping answers}: List all of the answers in [response] that also appear in [correct\_answer]. You can consider an answer from [response] to match with an answer in [correct\_answer] if it is equivalent or is within a small margin of error for numerical problems, a margin of 1 to 5.5 percentage points is acceptable. List all of the [response] answer appearing in [correct\_answer] with each answer delimited by `\#\#\#'. If the number of overlapping answers is zero, output `NULL'.\\

    
\bottomrule
\end{tabular}
\caption{The LLM-as-judge evaluation prompt.}
\label{fig:prompt_llm_judge_eval}
\end{figure*}

\begin{figure*}[t]\setlength{\belowcaptionskip}{-8pt}
\footnotesize
  \begin{tabular}{p{0.97\linewidth}}
    \toprule \textbf{Oracle retrieval:}\\
    *** Above are multiple excerpts of paragraphs and tables, each of document has a title, followed by the actual content. These documents contain highly relevant information which should help you answer the user question. \\

    \midrule \textbf{BM25 retrieval:}\\
    *** Above are provided with multiple excerpts of paragraphs and tables, each of document has a title, followed by the actual content. Some of these documents might contain helpful information to answering the question. In case that the information in the document is relevant, you may use it to solve the question. If a document is irrelevant feel free to ignore it when answering. \\
  
\bottomrule
\end{tabular}
\caption{Retrieval-augmented QA instructions which follow the same system prompt from Figure~\ref{fig:prompt_comlex_qa_no_cot}.}
\label{fig:prompt_retrieval_augmented}
\end{figure*}

\begin{figure*}[t]\setlength{\belowcaptionskip}{-8pt}
\footnotesize
  \begin{tabular}{p{0.97\linewidth}}
    \toprule \textbf{System prompt:}\\
    You are a helpful question answering assistant. Your task is to answer a complex question provided by the user. You may generate an explanation before providing the answer, as a chain-of-thought reasoning. The answer must be generated as a concise list of one or more entities, numbers or dates. You must always answer the question, even if your information is not up-to-date, please answer based on it. \\

Your response must use the following format: \\
Let’s think step by step: \{EXPLANATION\}\\
Answers: \{ANSWERS\}\\
Keep the explanation as concise as possible. ANSWERS is a list of potential answers, each answer in a separate line. You must end your response after the final answer. You must always answer the question, even if your information is not up-to-date, please answer based on it. \vspace{0.005mm}\\ 
    
    \midrule \textbf{Prompt examples:}\\
    \vspace{0.005mm}
    \underline{Question}: What are all the species of bears that currently exist in the world? \\
    \underline{Let’s think step by step}: To find the answer we will need to return all the different types of bears that are not gone extinct.\\
\underline{Answers}: 
American Black Bear,
Asian Black Bear,
Brown Bear,
Polar Bear,
Sloth Bear,
Spectacled Bear,
Sun Bear,
Giant Panda \\
    
    \vspace{0.005mm}

    \vspace{0.005mm}
    \underline{Question}: List all of the offices in Italy's current cabinet? \\
    \underline{Let’s think step by step}: To answer this question, we need to return all the ministerial positions of Italy's incumbent government.\\
\underline{Answers}: 
Prime Minister, 
Deputy Prime Minister, 
Minister of Foreign Affairs and International Cooperation, 
Minister of the Interior, 
Minister of Justice, 
Minister of Defence, 
Minister of Economy and Finance, 
Minister of Business and Made in Italy, 
Minister of Agriculture, Food Sovereignty and Forests, 
Minister for the Environment and Energy Security, 
Minister of Infrastructure and Transport, 
Minister of Labour and Social Policies, 
Minister of Education and Merit, 
Minister of University and Research, 
Minister of Culture, 
Minister of Health, 
Minister of Tourism, 
Minister for Relations with Parliament,
Minister for Public Administration,
Minister for Regional Affairs and Autonomies,
Minister for Civil Protection and Maritime Policies,
Minister for European Affairs, Southern Italy, Cohesion Policy and the NRRP,
Minister for Sport and Youth,
Minister for Family, Birth Rate and Equal Opportunities,
Minister for Disabilities,
Minister for Institutional Reforms and Regulatory Simplification,
Secretary of the Council of Ministers\\

    ...\\

\bottomrule
\end{tabular}
\caption{Excerpt of the list QA prompt, used to evaluate LLM performance on list questions.}
\label{fig:prompt_list_qa}
\end{figure*}

\begin{figure*}[t]\setlength{\belowcaptionskip}{-8pt}
\footnotesize
  \begin{tabular}{p{0.97\linewidth}}
   \toprule \textbf{System prompt:}\\
    You are a helpful question answering assistant. Your task is to answer a complex question provided by the user. You may generate an explanation before providing the answer, as a chain-of-thought reasoning. The answer must be generated as a concise list of one or more entities, numbers or dates. You must always answer the question, even if your information is not up-to-date, please answer based on it. \\

    Your response must use the following format: \\
    Let’s think step by step: \{EXPLANATION\}\\
    Answers: \{ANSWERS\}\\
    Keep the explanation as concise as possible. ANSWERS is a list of potential answers, each answer in a separate line. You must end your response after the final answer. You must always answer the question, even if your information is not up-to-date, please answer based on it. \vspace{0.005mm}\\ 
    
    \midrule \textbf{Prompt examples:}\\
    \vspace{0.005mm}
    \underline{Question}: Is Toosie Slide a musical collaboration? \\
    \underline{Let’s think step by step}: "Toosie Slide" is not a musical collaboration. It is a solo song by the Canadian rapper Drake. Released on April 3, 2020, the song became popular on social media platforms, especially TikTok, due to its catchy hook and dance challenge. The dance associated with the song was created by the dancer Toosie, which is where the song gets its name. However, there are no other artists featured in the track.\\
\underline{Answers}: no \\

    ... \\

\bottomrule
\end{tabular}
\caption{Excerpt of the factoid QA prompt, used for factoid QA validation.}
\label{fig:prompt_factoid_qa}
\end{figure*}

\begin{figure*}[t]\setlength{\belowcaptionskip}{-8pt}
\footnotesize
  \begin{tabular}{p{0.97\linewidth}}
    \toprule \textbf{System prompt:}\\
    You are a helpful assistant, tasked with determining whether two answers are in agreement with each other. Given a question and two potential answers, determine whether both answers are the same. If both answers refer to the same entity (e.g. `US' and `United States of America') or person (`President Biden' and `Joe Bide') In the case of numeric answers, we regard numbers that are very close to each other as being similar. Your response should be either `yes' or `no'.\vspace{0.005mm} \\
    
    \midrule \textbf{Prompt examples:}\\
    \vspace{0.005mm}
    \underline{Question}: For how long did Fahd serve as king of Saudi Arabia? \\
    \underline{Answer 1}: 23 years \\
    \underline{Answer 2}: 13 June 1982 – 1 August 2005\\
    \underline{Are these answers the same?} no \\

\bottomrule
\end{tabular}
\caption{Excerpt of the answer agreement prompt, used for factoid QA validation.}
\label{fig:prompt_answer_agreement}
\end{figure*}

\begin{figure*}[t]\setlength{\belowcaptionskip}{-8pt}
\footnotesize
  \begin{tabular}{p{0.97\linewidth}}
    \toprule \textbf{System prompt:}\\
    You are a helpful question answering assistant. Your task is to answer a complex question provided by the user. You may generate an explanation before providing the answer, as a chain-of-thought reasoning. The answer must be generated as a concise list of one or more entities, numbers or dates. You must always answer the question, even if your information is not up-to-date, please answer based on it. 
Your response must use the following format: 

Let’s think step by step: \{EXPLANATION\}

Answers: \{ANSWERS\}

Keep the explanation as concise as possible. ANSWERS is a list of potential answers, separated by commas. You must end your response after the final answer. You must always answer the question, even if your information is not up-to-date, please answer based on it.\vspace{0.005mm} \\
    
    \midrule \textbf{Prompt examples:}\\
    \vspace{0.005mm}
\underline{Question}: How many plays did Arthur Miller write between the ages of 20 and 40? \\
    \underline{Let’s think step by step}: 
    
    to answer the question we need to find out what were the years in which Arthur Miller turned 20 and 40. Then, we need to review the plays written by Miller and return only the plays that were written between those years.
\\
    \underline{Answers}: 15\\
    


\bottomrule
\end{tabular}
\caption{Excerpt of the few-shot chain-of-thought prompt (CoT).}
\label{fig:prompt_cot}
\end{figure*}

\begin{figure*}[t]\setlength{\belowcaptionskip}{-8pt}
\footnotesize
  \begin{tabular}{p{0.97\linewidth}}
    \toprule \textbf{System prompt:}\\
    You are a helpful question answering assistant. Your task is to answer a complex question provided by the user. You may generate an explanation before providing the answer, as a chain-of-thought reasoning. The answer must be generated as a concise list of one or more entities, numbers or dates. You must always answer the question, even if your information is not up-to-date, please answer based on it. 
Your response must use the following format: 

Let’s think step by step: \{EXPLANATION\}

Answers: \{ANSWERS\}

Keep the explanation as concise as possible. ANSWERS is a list of potential answers, separated by commas. You must end your response after the final answer. You must always answer the question, even if your information is not up-to-date, please answer based on it.\vspace{0.005mm} \\
    
    \midrule \textbf{Prompt examples:}\\
    \vspace{0.005mm}
\underline{Question}: How many plays did Arthur Miller write between the ages of 20 and 40? \\
    \underline{Let’s think step by step}: 
    
    1. Arthur Miller was born on October 17, 1915, so he turned 20 on October 17, 1935 and he turned 40 on October 17, 1955.
    
    2. Between October 17, 1935 and October 17, 1955 Arthur Miller wrote the following plays: No Villain (1936), They Too Arise (1937), Honors at Dawn (1938), The Grass Still Grows (1938), The Great Disobedience (1938), Listen My Children (1939), The Golden Years (1940), The Half-Bridge (1943), The Man Who Had All the Luck (1944), All My Sons (1947), Death of a Salesman (1949), An Enemy of the People (1950), The Crucible (1953), A View from the Bridge (1955), A Memory of Two Mondays (1955).
    
    3. Therefore Miller had written 15 plays between October 17, 1935 and October 17, 1955.
\\
    \underline{Answers}: 15\\
    
    \vspace{0.005mm}

  \\
\bottomrule
\end{tabular}
\caption{Excerpt of the few-shot chain-of-thought prompt which includes intermediate answers (CoT+Ans).}
\label{fig:prompt_cot_answers}
\end{figure*}

%% file: main-9013-Wolfson.bbl
\begin{thebibliography}{71}
\expandafter\ifx\csname natexlab\endcsname\relax\def\natexlab#1{#1}\fi

\bibitem[{Abujabal et~al.(2019)Abujabal, Roy, Yahya, and Weikum}]{Abujabal2018ComQAAC}
Abdalghani Abujabal, Rishiraj~Saha Roy, Mohamed Yahya, and Gerhard Weikum. 2019.
\newblock Comqa: A community-sourced dataset for complex factoid question answering with paraphrase clusters.
\newblock In \emph{North American Association for Computational Linguistics (NAACL)}.

\bibitem[{Amouyal et~al.(2023)Amouyal, Wolfson, Rubin, Yoran, Herzig, and Berant}]{amouyal-etal-2023-qampari}
Samuel Amouyal, Tomer Wolfson, Ohad Rubin, Ori Yoran, Jonathan Herzig, and Jonathan Berant. 2023.
\newblock \href {https://aclanthology.org/2023.gem-1.9} {{QAMPARI}: A benchmark for open-domain questions with many answers}.
\newblock In \emph{Proceedings of the Third Workshop on Natural Language Generation, Evaluation, and Metrics (GEM)}, pages 97--110, Singapore. Association for Computational Linguistics.

\bibitem[{Anthropic(2025)}]{AnthropicClaudeFour}
Anthropic. 2025.
\newblock \href {https://www-cdn.anthropic.com/6d8a8055020700718b0c49369f60816ba2a7c285.pdf} {System card: Claude opus 4 \& claude sonnet 4}.

\bibitem[{Asai et~al.(2024)Asai, He*, Shao*, Shi, Singh, Chang, Lo, Soldaini, Feldman, Mike, Wadden, Latzke, Minyang, Ji, Liu, Tong, Wu, Xiong, Zettlemoyer, Weld, Neubig, Downey, Yih, Koh, and Hajishirzi}]{openscholar}
Akari Asai, Jacqueline He*, Rulin Shao*, Weijia Shi, Amanpreet Singh, Joseph~Chee Chang, Kyle Lo, Luca Soldaini, Sergey Feldman, Tian, D’arcy Mike, David Wadden, Matt Latzke, Minyang, Pan Ji, Shengyan Liu, Hao Tong, Bohao Wu, Yanyu Xiong, Luke Zettlemoyer, Dan Weld, Graham Neubig, Doug Downey, Wen-tau Yih, Pang~Wei Koh, and Hannaneh Hajishirzi. 2024.
\newblock {OpenScholar}: Synthesizing scientific literature with retrieval-augmented language models.
\newblock \emph{Arxiv}.

\bibitem[{Asai et~al.(2023)Asai, Wu, Wang, Sil, and Hajishirzi}]{asai2023selfrag}
Akari Asai, Zeqiu Wu, Yizhong Wang, Avirup Sil, and Hannaneh Hajishirzi. 2023.
\newblock \href {https://arxiv.org/abs/2310.11511} {{Self-RAG}: Learning to retrieve, generate, and critique through self-reflection}.
\newblock \emph{arXiv preprint arXiv:2310.11511}.

\bibitem[{Bajaj et~al.(2016)Bajaj, Campos, Craswell, Deng, Gao, Liu, Majumder, McNamara, Mitra, Nguyen, Rosenberg, Song, Stoica, Tiwary, and Wang}]{nguyen2016ms}
Payal Bajaj, Daniel Campos, Nick Craswell, Li~Deng, Jianfeng Gao, Xiaodong Liu, Rangan Majumder, Andrew McNamara, Bhaskar Mitra, Tri Nguyen, Mir Rosenberg, Xia Song, Alina Stoica, Saurabh Tiwary, and Tong Wang. 2016.
\newblock \href {https://arxiv.org/abs/1611.09268} {{MS MARCO}: A human generated {MA}chine {R}eading {CO}mprehension dataset}.
\newblock \emph{ArXiv}, abs/1611.09268.

\bibitem[{Bohnet et~al.(2022)Bohnet, Tran, Verga, Aharoni, Andor, Soares, Eisenstein, Ganchev, Herzig, Hui, Kwiatkowski, Ma, Ni, Schuster, Cohen, Collins, Das, Metzler, Petrov, and Webster}]{Bohnet2022AttributedQA}
Bernd Bohnet, Vinh~Q. Tran, Pat Verga, Roee Aharoni, Daniel Andor, Livio~Baldini Soares, Jacob Eisenstein, Kuzman Ganchev, Jonathan Herzig, Kai Hui, Tom Kwiatkowski, Ji~Ma, Jianmo Ni, Tal Schuster, William~W. Cohen, Michael Collins, Dipanjan Das, Donald Metzler, Slav Petrov, and Kellie Webster. 2022.
\newblock \href {https://api.semanticscholar.org/CorpusID:254685584} {Attributed question answering: Evaluation and modeling for attributed large language models}.
\newblock \emph{ArXiv}, abs/2212.08037.

\bibitem[{Bowman and Dahl(2021)}]{bowman-dahl-2021-will}
Samuel~R. Bowman and George Dahl. 2021.
\newblock \href {https://doi.org/10.18653/v1/2021.naacl-main.385} {What will it take to fix benchmarking in natural language understanding?}
\newblock In \emph{Proceedings of the 2021 Conference of the North American Chapter of the Association for Computational Linguistics: Human Language Technologies}, pages 4843--4855, Online. Association for Computational Linguistics.

\bibitem[{Clark et~al.(2019)Clark, Lee, Chang, Kwiatkowski, Collins, and Toutanova}]{clark-etal-2019-boolq}
Christopher Clark, Kenton Lee, Ming-Wei Chang, Tom Kwiatkowski, Michael Collins, and Kristina Toutanova. 2019.
\newblock \href {https://doi.org/10.18653/v1/N19-1300} {{B}ool{Q}: Exploring the surprising difficulty of natural yes/no questions}.
\newblock In \emph{Proceedings of the 2019 Conference of the North {A}merican Chapter of the Association for Computational Linguistics: Human Language Technologies, Volume 1 (Long and Short Papers)}, pages 2924--2936, Minneapolis, Minnesota. Association for Computational Linguistics.

\bibitem[{Dasigi et~al.(2021)Dasigi, Lo, Beltagy, Cohan, Smith, and Gardner}]{dasigi-etal-2021-dataset}
Pradeep Dasigi, Kyle Lo, Iz~Beltagy, Arman Cohan, Noah~A. Smith, and Matt Gardner. 2021.
\newblock \href {https://doi.org/10.18653/v1/2021.naacl-main.365} {A dataset of information-seeking questions and answers anchored in research papers}.
\newblock In \emph{Proceedings of the 2021 Conference of the North American Chapter of the Association for Computational Linguistics: Human Language Technologies}, pages 4599--4610, Online. Association for Computational Linguistics.

\bibitem[{{DeepSeek AI}(2025)}]{deepseekai2025deepseekr1incentivizingreasoningcapability}
{DeepSeek AI}. 2025.
\newblock {DeepSeek-R1}: Incentivizing reasoning capability in {LLMs} via reinforcement learning.
\newblock \emph{arXiv}, abs/2501.12948.

\bibitem[{Dua et~al.(2019)Dua, Wang, Dasigi, Stanovsky, Singh, and Gardner}]{dua-etal-2019-drop}
Dheeru Dua, Yizhong Wang, Pradeep Dasigi, Gabriel Stanovsky, Sameer Singh, and Matt Gardner. 2019.
\newblock \href {https://doi.org/10.18653/v1/N19-1246} {{DROP}: A reading comprehension benchmark requiring discrete reasoning over paragraphs}.
\newblock In \emph{Proceedings of the 2019 Conference of the North {A}merican Chapter of the Association for Computational Linguistics: Human Language Technologies, Volume 1 (Long and Short Papers)}, pages 2368--2378, Minneapolis, Minnesota. Association for Computational Linguistics.

\bibitem[{Gao et~al.(2023)Gao, Yen, Yu, and Chen}]{gao-etal-2023-enabling}
Tianyu Gao, Howard Yen, Jiatong Yu, and Danqi Chen. 2023.
\newblock \href {https://doi.org/10.18653/v1/2023.emnlp-main.398} {Enabling large language models to generate text with citations}.
\newblock In \emph{Proceedings of the 2023 Conference on Empirical Methods in Natural Language Processing}, pages 6465--6488, Singapore. Association for Computational Linguistics.

\bibitem[{Gemini~Team(2025)}]{Comanici2025Gemini2P}
Google Gemini~Team. 2025.
\newblock \href {https://api.semanticscholar.org/CorpusID:280151524} {Gemini 2.5: Pushing the frontier with advanced reasoning, multimodality, long context, and next generation agentic capabilities}.
\newblock \emph{ArXiv}, abs/2507.06261.

\bibitem[{Geva et~al.(2021)Geva, Khashabi, Segal, Khot, Roth, and Berant}]{geva-etal-2021-aristotle}
Mor Geva, Daniel Khashabi, Elad Segal, Tushar Khot, Dan Roth, and Jonathan Berant. 2021.
\newblock \href {https://doi.org/10.1162/tacl_a_00370} {Did aristotle use a laptop? a question answering benchmark with implicit reasoning strategies}.
\newblock \emph{Transactions of the Association for Computational Linguistics}, 9:346--361.

\bibitem[{Geva et~al.(2022)Geva, Wolfson, and Berant}]{geva-etal-2022-break}
Mor Geva, Tomer Wolfson, and Jonathan Berant. 2022.
\newblock \href {https://doi.org/10.1162/tacl_a_00450} {Break, perturb, build: Automatic perturbation of reasoning paths through question decomposition}.
\newblock \emph{Transactions of the Association for Computational Linguistics}, 10:111--126.

\bibitem[{Guo et~al.(2024)Guo, Chen, Wang, Chang, Pei, Chawla, Wiest, and Zhang}]{Guo2024LargeLM}
Taicheng Guo, Xiuying Chen, Yaqi Wang, Ruidi Chang, Shichao Pei, N.~Chawla, Olaf Wiest, and Xiangliang Zhang. 2024.
\newblock \href {https://api.semanticscholar.org/CorpusID:267412980} {Large language model based multi-agents: A survey of progress and challenges}.
\newblock In \emph{International Joint Conference on Artificial Intelligence}.

\bibitem[{Hongjin et~al.(2025)Hongjin, Yen, Xia, Shi, Muennighoff, Wang, Haisu, Shi, Siegel, Tang et~al.}]{hongjin2024bright}
SU~Hongjin, Howard Yen, Mengzhou Xia, Weijia Shi, Niklas Muennighoff, Han-yu Wang, Liu Haisu, Quan Shi, Zachary~S Siegel, Michael Tang, et~al. 2025.
\newblock Bright: A realistic and challenging benchmark for reasoning-intensive retrieval.
\newblock In \emph{The International Conference on Learning Representations}.

\bibitem[{Hsieh et~al.(2024)Hsieh, Sun, Kriman, Acharya, Rekesh, Jia, Zhang, and Ginsburg}]{hsieh2024ruler}
Cheng-Ping Hsieh, Simeng Sun, Samuel Kriman, Shantanu Acharya, Dima Rekesh, Fei Jia, Yang Zhang, and Boris Ginsburg. 2024.
\newblock Ruler: What's the real context size of your long-context language models?
\newblock \emph{arXiv preprint arXiv:2404.06654}.

\bibitem[{Huang et~al.(2025)Huang, Chen, Zhang, Li, Fang, Yang, Li, Shang, Xu, Hao, Shao, and Wang}]{Huang2025DeepRA}
Yuxuan Huang, Yihang Chen, Haozheng Zhang, Kang Li, Meng Fang, Linyi Yang, Xiaoguang Li, Lifeng Shang, Songcen Xu, Jianye Hao, Kun Shao, and Jun Wang. 2025.
\newblock \href {https://api.semanticscholar.org/CorpusID:279999119} {Deep research agents: A systematic examination and roadmap}.
\newblock \emph{ArXiv}, abs/2506.18096.

\bibitem[{Jacovi et~al.(2025)Jacovi, Wang, Alberti, Tao, Lipovetz, Olszewska, Haas, Liu, Keating, Bloniarz, Saroufim, Fry, Marcus, Kukliansky, Tomar, Swirhun, Xing, Wang, Gurumurthy, Aaron, Ambar, Fellinger, Wang, Zhang, Goldshtein, and Das}]{jacovi2025factsgroundingleaderboardbenchmarking}
Alon Jacovi, Andrew Wang, Chris Alberti, Connie Tao, Jon Lipovetz, Kate Olszewska, Lukas Haas, Michelle Liu, Nate Keating, Adam Bloniarz, Carl Saroufim, Corey Fry, Dror Marcus, Doron Kukliansky, Gaurav~Singh Tomar, James Swirhun, Jinwei Xing, Lily Wang, Madhu Gurumurthy, Michael Aaron, Moran Ambar, Rachana Fellinger, Rui Wang, Zizhao Zhang, Sasha Goldshtein, and Dipanjan Das. 2025.
\newblock \href {http://arxiv.org/abs/2501.03200} {The facts grounding leaderboard: Benchmarking llms' ability to ground responses to long-form input}.

\bibitem[{Jin et~al.(2019)Jin, Dhingra, Liu, Cohen, and Lu}]{jin-etal-2019-pubmedqa}
Qiao Jin, Bhuwan Dhingra, Zhengping Liu, William Cohen, and Xinghua Lu. 2019.
\newblock \href {https://doi.org/10.18653/v1/D19-1259} {{P}ub{M}ed{QA}: A dataset for biomedical research question answering}.
\newblock In \emph{Proceedings of the 2019 Conference on Empirical Methods in Natural Language Processing and the 9th International Joint Conference on Natural Language Processing (EMNLP-IJCNLP)}, pages 2567--2577, Hong Kong, China. Association for Computational Linguistics.

\bibitem[{Kamalloo et~al.(2023)Kamalloo, Dziri, Clarke, and Rafiei}]{kamalloo-etal-2023-evaluating}
Ehsan Kamalloo, Nouha Dziri, Charles Clarke, and Davood Rafiei. 2023.
\newblock \href {https://doi.org/10.18653/v1/2023.acl-long.307} {Evaluating open-domain question answering in the era of large language models}.
\newblock In \emph{Proceedings of the 61st Annual Meeting of the Association for Computational Linguistics (Volume 1: Long Papers)}, pages 5591--5606, Toronto, Canada. Association for Computational Linguistics.

\bibitem[{Katz et~al.(2023)Katz, Vetzler, Cohen, and Goldberg}]{katz-etal-2023-neretrieve}
Uri Katz, Matan Vetzler, Amir Cohen, and Yoav Goldberg. 2023.
\newblock \href {https://doi.org/10.18653/v1/2023.findings-emnlp.218} {{NER}etrieve: Dataset for next generation named entity recognition and retrieval}.
\newblock In \emph{Findings of the Association for Computational Linguistics: EMNLP 2023}, pages 3340--3354, Singapore. Association for Computational Linguistics.

\bibitem[{Kojima et~al.(2022)Kojima, Gu, Reid, Matsuo, and Iwasawa}]{kojima2022large}
Takeshi Kojima, Shixiang~Shane Gu, Machel Reid, Yutaka Matsuo, and Yusuke Iwasawa. 2022.
\newblock \href {https://openreview.net/forum?id=6p3AuaHAFiN} {Large language models are zero-shot reasoners}.
\newblock In \emph{ICML 2022 Workshop on Knowledge Retrieval and Language Models}.

\bibitem[{Krishna et~al.(2024)Krishna, Krishna, Mohananey, Schwarcz, Stambler, Upadhyay, and Faruqui}]{krishna2024factfetchreasonunified}
Satyapriya Krishna, Kalpesh Krishna, Anhad Mohananey, Steven Schwarcz, Adam Stambler, Shyam Upadhyay, and Manaal Faruqui. 2024.
\newblock \href {http://arxiv.org/abs/2409.12941} {Fact, fetch, and reason: A unified evaluation of retrieval-augmented generation}.

\bibitem[{Krithara et~al.(2023)Krithara, Nentidis, Bougiatiotis, and Paliouras}]{krithara2023bioasq}
Anastasia Krithara, Anastasios Nentidis, Konstantinos Bougiatiotis, and Georgios Paliouras. 2023.
\newblock Bioasq-qa: A manually curated corpus for biomedical question answering.
\newblock \emph{Scientific Data}, 10(1):170.

\bibitem[{Kwiatkowski et~al.(2019)Kwiatkowski, Palomaki, Redfield, Collins, Parikh, Alberti, Epstein, Polosukhin, Devlin, Lee, Toutanova, Jones, Kelcey, Chang, Dai, Uszkoreit, Le, and Petrov}]{kwiatkowski-etal-2019-natural}
Tom Kwiatkowski, Jennimaria Palomaki, Olivia Redfield, Michael Collins, Ankur Parikh, Chris Alberti, Danielle Epstein, Illia Polosukhin, Jacob Devlin, Kenton Lee, Kristina Toutanova, Llion Jones, Matthew Kelcey, Ming-Wei Chang, Andrew~M. Dai, Jakob Uszkoreit, Quoc Le, and Slav Petrov. 2019.
\newblock \href {https://doi.org/10.1162/tacl_a_00276} {Natural questions: A benchmark for question answering research}.
\newblock \emph{Transactions of the Association for Computational Linguistics}, 7:452--466.

\bibitem[{Lewis et~al.(2020)Lewis, Perez, Piktus, Petroni, Karpukhin, Goyal, K{\"u}ttler, Lewis, Yih, Rockt{\"a}schel et~al.}]{lewis2020retrieval}
Patrick Lewis, Ethan Perez, Aleksandra Piktus, Fabio Petroni, Vladimir Karpukhin, Naman Goyal, Heinrich K{\"u}ttler, Mike Lewis, Wen-tau Yih, Tim Rockt{\"a}schel, et~al. 2020.
\newblock Retrieval-augmented generation for knowledge-intensive nlp tasks.
\newblock \emph{Advances in Neural Information Processing Systems}, 33:9459--9474.

\bibitem[{Li et~al.(2024)Li, Zhou, Wang, Fu, Roth, and Chen}]{li-etal-2024-deceptive}
Bangzheng Li, Ben Zhou, Fei Wang, Xingyu Fu, Dan Roth, and Muhao Chen. 2024.
\newblock \href {https://doi.org/10.18653/v1/2024.naacl-long.424} {Deceptive semantic shortcuts on reasoning chains: How far can models go without hallucination?}
\newblock In \emph{Proceedings of the 2024 Conference of the North American Chapter of the Association for Computational Linguistics: Human Language Technologies (Volume 1: Long Papers)}, pages 7675--7688, Mexico City, Mexico. Association for Computational Linguistics.

\bibitem[{Lin et~al.(2024)Lin, Deng, Chandu, Brahman, Ravichander, Pyatkin, Dziri, Bras, and Choi}]{lin2024wildbench}
Bill~Yuchen Lin, Yuntian Deng, Khyathi Chandu, Faeze Brahman, Abhilasha Ravichander, Valentina Pyatkin, Nouha Dziri, Ronan~Le Bras, and Yejin Choi. 2024.
\newblock \href {http://arxiv.org/abs/2406.04770} {Wildbench: Benchmarking llms with challenging tasks from real users in the wild}.

\bibitem[{Malaviya et~al.(2024)Malaviya, Lee, Chen, Sieber, Yatskar, and Roth}]{malaviya-etal-2024-expertqa}
Chaitanya Malaviya, Subin Lee, Sihao Chen, Elizabeth Sieber, Mark Yatskar, and Dan Roth. 2024.
\newblock \href {https://doi.org/10.18653/v1/2024.naacl-long.167} {{E}xpert{QA}: Expert-curated questions and attributed answers}.
\newblock In \emph{Proceedings of the 2024 Conference of the North American Chapter of the Association for Computational Linguistics: Human Language Technologies (Volume 1: Long Papers)}, pages 3025--3045, Mexico City, Mexico. Association for Computational Linguistics.

\bibitem[{Malaviya et~al.(2023)Malaviya, Shaw, Chang, Lee, and Toutanova}]{malaviya-etal-2023-quest}
Chaitanya Malaviya, Peter Shaw, Ming-Wei Chang, Kenton Lee, and Kristina Toutanova. 2023.
\newblock \href {https://doi.org/10.18653/v1/2023.acl-long.784} {{QUEST}: A retrieval dataset of entity-seeking queries with implicit set operations}.
\newblock In \emph{Proceedings of the 61st Annual Meeting of the Association for Computational Linguistics (Volume 1: Long Papers)}, pages 14032--14047, Toronto, Canada. Association for Computational Linguistics.

\bibitem[{OpenAI(2024)}]{openai2024gpt4technicalreport}
OpenAI. 2024.
\newblock \href {http://arxiv.org/abs/2303.08774} {{GPT-4} technical report}.

\bibitem[{{OpenAI}(2024)}]{openai2024openaio1card}
{OpenAI}. 2024.
\newblock \href {http://arxiv.org/abs/2412.16720} {{OpenAI} {o1} system card}.

\bibitem[{OpenAI(2025{\natexlab{a}})}]{OpenAIDeepRS}
OpenAI. 2025{\natexlab{a}}.
\newblock \href {https://api.semanticscholar.org/CorpusID:276648957} {Deep research system card}.

\bibitem[{OpenAI(2025{\natexlab{b}})}]{OpenAIGPTFive}
OpenAI. 2025{\natexlab{b}}.
\newblock \href {https://cdn.openai.com/gpt-5-system-card.pdf} {Gpt-5 system card}.

\bibitem[{Phan et~al.(2025)Phan, Gatti, Han, Li, Hu, Zhang, Zhang, Shaaban, Ling, Shi, Choi, Agrawal, Chopra, Khoja, Kim, Ren, Hausenloy, Zhang, Maz, Yue, Wang, and Hendrycks}]{phan2025humanitysexam}
Long Phan, Alice Gatti, Ziwen Han, Nathaniel Li, Josephina Hu, Hugh Zhang, Chen Bo~Calvin Zhang, Mohamed Shaaban, John Ling, Sean Shi, Michael Choi, Anish Agrawal, Arnav Chopra, Adam Khoja, Ryan Kim, Richard Ren, Jason Hausenloy, Oliver Zhang, Mantas Maz, Summer Yue, Alexandr Wang, and Dan Hendrycks. 2025.
\newblock \href {http://arxiv.org/abs/2501.14249} {Humanity's last exam}.

\bibitem[{Press et~al.(2023)Press, Zhang, Min, Schmidt, Smith, and Lewis}]{press-etal-2023-measuring}
Ofir Press, Muru Zhang, Sewon Min, Ludwig Schmidt, Noah Smith, and Mike Lewis. 2023.
\newblock \href {https://doi.org/10.18653/v1/2023.findings-emnlp.378} {Measuring and narrowing the compositionality gap in language models}.
\newblock In \emph{Findings of the Association for Computational Linguistics: EMNLP 2023}, pages 5687--5711, Singapore. Association for Computational Linguistics.

\bibitem[{Rajpurkar et~al.(2016)Rajpurkar, Zhang, Lopyrev, and Liang}]{rajpurkar-etal-2016-squad}
Pranav Rajpurkar, Jian Zhang, Konstantin Lopyrev, and Percy Liang. 2016.
\newblock \href {https://doi.org/10.18653/v1/D16-1264} {{SQ}u{AD}: 100,000+ questions for machine comprehension of text}.
\newblock In \emph{Proceedings of the 2016 Conference on Empirical Methods in Natural Language Processing}, pages 2383--2392, Austin, Texas. Association for Computational Linguistics.

\bibitem[{Ram et~al.(2023)Ram, Levine, Dalmedigos, Muhlgay, Shashua, Leyton-Brown, and Shoham}]{ram-etal-2023-context}
Ori Ram, Yoav Levine, Itay Dalmedigos, Dor Muhlgay, Amnon Shashua, Kevin Leyton-Brown, and Yoav Shoham. 2023.
\newblock \href {https://doi.org/10.1162/tacl_a_00605} {In-context retrieval-augmented language models}.
\newblock \emph{Transactions of the Association for Computational Linguistics}, 11:1316--1331.

\bibitem[{Rassin et~al.(2024)Rassin, Fairstein, Kalinsky, Kushilevitz, Cohen, Libov, and Goldberg}]{rassin-etal-2024-evaluating}
Royi Rassin, Yaron Fairstein, Oren Kalinsky, Guy Kushilevitz, Nachshon Cohen, Alexander Libov, and Yoav Goldberg. 2024.
\newblock \href {https://doi.org/10.18653/v1/2024.emnlp-main.171} {Evaluating {D}-{MERIT} of partial-annotation on information retrieval}.
\newblock In \emph{Proceedings of the 2024 Conference on Empirical Methods in Natural Language Processing}, pages 2913--2932, Miami, Florida, USA. Association for Computational Linguistics.

\bibitem[{Robertson et~al.(2009)Robertson, Zaragoza et~al.}]{robertson2009probabilistic}
Stephen Robertson, Hugo Zaragoza, et~al. 2009.
\newblock The probabilistic relevance framework: Bm25 and beyond.
\newblock \emph{Foundations and Trends{\textregistered} in Information Retrieval}, 3(4):333--389.

\bibitem[{Rodriguez and Boyd-Graber(2021)}]{rodriguez-boyd-graber-2021-evaluation}
Pedro Rodriguez and Jordan Boyd-Graber. 2021.
\newblock \href {https://doi.org/10.18653/v1/2021.emnlp-main.758} {Evaluation paradigms in question answering}.
\newblock In \emph{Proceedings of the 2021 Conference on Empirical Methods in Natural Language Processing}, pages 9630--9642, Online and Punta Cana, Dominican Republic. Association for Computational Linguistics.

\bibitem[{Saparina and Osokin(2021)}]{saparina-osokin-2021-sparqling}
Irina Saparina and Anton Osokin. 2021.
\newblock \href {https://doi.org/10.18653/v1/2021.emnlp-main.708} {{SPARQL}ing database queries from intermediate question decompositions}.
\newblock In \emph{Proceedings of the 2021 Conference on Empirical Methods in Natural Language Processing}, pages 8984--8998, Online and Punta Cana, Dominican Republic. Association for Computational Linguistics.

\bibitem[{Scholman et~al.(2022)Scholman, Pyatkin, Yung, Dagan, Tsarfaty, and Demberg}]{scholman-etal-2022-design}
Merel Scholman, Valentina Pyatkin, Frances Yung, Ido Dagan, Reut Tsarfaty, and Vera Demberg. 2022.
\newblock \href {https://aclanthology.org/2022.lrec-1.231} {Design choices in crowdsourcing discourse relation annotations: The effect of worker selection and training}.
\newblock In \emph{Proceedings of the Thirteenth Language Resources and Evaluation Conference}, pages 2148--2156, Marseille, France. European Language Resources Association.

\bibitem[{Sen et~al.(2022)Sen, Aji, and Saffari}]{sen-etal-2022-mintaka}
Priyanka Sen, Alham~Fikri Aji, and Amir Saffari. 2022.
\newblock \href {https://aclanthology.org/2022.coling-1.138} {Mintaka: A complex, natural, and multilingual dataset for end-to-end question answering}.
\newblock In \emph{Proceedings of the 29th International Conference on Computational Linguistics}, pages 1604--1619, Gyeongju, Republic of Korea. International Committee on Computational Linguistics.

\bibitem[{Shannon(1948)}]{shannon1948mathematical}
Claude~Elwood Shannon. 1948.
\newblock A mathematical theory of communication.
\newblock \emph{The Bell system technical journal}, 27(3):379--423.

\bibitem[{Talmor et~al.(2021)Talmor, Yoran, Catav, Lahav, Wang, Asai, Ilharco, Hajishirzi, and Berant}]{talmor2021multimodalqa}
Alon Talmor, Ori Yoran, Amnon Catav, Dan Lahav, Yizhong Wang, Akari Asai, Gabriel Ilharco, Hannaneh Hajishirzi, and Jonathan Berant. 2021.
\newblock \href {https://openreview.net/forum?id=ee6W5UgQLa} {Multimodal{\{}qa{\}}: complex question answering over text, tables and images}.
\newblock In \emph{International Conference on Learning Representations}.

\bibitem[{Thakur et~al.(2021)Thakur, Reimers, R{\"u}ckl{\'e}, Srivastava, and Gurevych}]{thakur2021beir}
Nandan Thakur, Nils Reimers, Andreas R{\"u}ckl{\'e}, Abhishek Srivastava, and Iryna Gurevych. 2021.
\newblock \href {https://openreview.net/forum?id=wCu6T5xFjeJ} {{BEIR}: A heterogeneous benchmark for zero-shot evaluation of information retrieval models}.
\newblock In \emph{Thirty-fifth Conference on Neural Information Processing Systems Datasets and Benchmarks Track (Round 2)}.

\bibitem[{Trivedi et~al.(2022)Trivedi, Balasubramanian, Khot, and Sabharwal}]{trivedi-etal-2022-musique}
Harsh Trivedi, Niranjan Balasubramanian, Tushar Khot, and Ashish Sabharwal. 2022.
\newblock \href {https://doi.org/10.1162/tacl_a_00475} {{M}u{S}i{Q}ue: Multihop questions via single-hop question composition}.
\newblock \emph{Transactions of the Association for Computational Linguistics}, 10:539--554.

\bibitem[{Trivedi et~al.(2023)Trivedi, Balasubramanian, Khot, and Sabharwal}]{trivedi-etal-2023-interleaving}
Harsh Trivedi, Niranjan Balasubramanian, Tushar Khot, and Ashish Sabharwal. 2023.
\newblock \href {https://doi.org/10.18653/v1/2023.acl-long.557} {Interleaving retrieval with chain-of-thought reasoning for knowledge-intensive multi-step questions}.
\newblock In \emph{Proceedings of the 61st Annual Meeting of the Association for Computational Linguistics (Volume 1: Long Papers)}, pages 10014--10037, Toronto, Canada. Association for Computational Linguistics.

\bibitem[{Voorhees and Tice(2000)}]{voorhees2000building}
Ellen~M Voorhees and Dawn~M Tice. 2000.
\newblock Building a question answering test collection.
\newblock In \emph{Proceedings of the 23rd annual international ACM SIGIR conference on Research and development in information retrieval}, pages 200--207.

\bibitem[{de~Vries et~al.(2020)de~Vries, Bahdanau, and Manning}]{devries2020ecologicallyvalidresearchlanguage}
Harm de~Vries, Dzmitry Bahdanau, and Christopher Manning. 2020.
\newblock \href {http://arxiv.org/abs/2007.14435} {Towards ecologically valid research on language user interfaces}.

\bibitem[{Vu et~al.(2024)Vu, Iyyer, Wang, Constant, Wei, Wei, Tar, Sung, Zhou, Le, and Luong}]{vu-etal-2024-freshllms}
Tu~Vu, Mohit Iyyer, Xuezhi Wang, Noah Constant, Jerry Wei, Jason Wei, Chris Tar, Yun-Hsuan Sung, Denny Zhou, Quoc Le, and Thang Luong. 2024.
\newblock \href {https://doi.org/10.18653/v1/2024.findings-acl.813} {{F}resh{LLM}s: Refreshing large language models with search engine augmentation}.
\newblock In \emph{Findings of the Association for Computational Linguistics: ACL 2024}, pages 13697--13720, Bangkok, Thailand. Association for Computational Linguistics.

\bibitem[{Wang et~al.(2024)Wang, Cheng, Guo, Yue, Ding, Xu, Wang, Hu, Zhang, and Zhang}]{wang2024evaluating}
Cunxiang Wang, Sirui Cheng, Qipeng Guo, Yuanhao Yue, Bowen Ding, Zhikun Xu, Yidong Wang, Xiangkun Hu, Zheng Zhang, and Yue Zhang. 2024.
\newblock Evaluating open-qa evaluation.
\newblock \emph{Advances in Neural Information Processing Systems}, 36.

\bibitem[{Wei et~al.(2024)Wei, Karina, Chung, Jiao, Papay, Glaese, Schulman, and Fedus}]{wei2024measuringshortformfactualitylarge}
Jason Wei, Nguyen Karina, Hyung~Won Chung, Yunxin~Joy Jiao, Spencer Papay, Amelia Glaese, John Schulman, and William Fedus. 2024.
\newblock \href {http://arxiv.org/abs/2411.04368} {Measuring short-form factuality in large language models}.

\bibitem[{Wei et~al.(2025)Wei, Sun, Papay, McKinney, Han, Fulford, Chung, Passos, Fedus, and Glaese}]{wei2025browsecompsimplechallengingbenchmark}
Jason Wei, Zhiqing Sun, Spencer Papay, Scott McKinney, Jeffrey Han, Isa Fulford, Hyung~Won Chung, Alex~Tachard Passos, William Fedus, and Amelia Glaese. 2025.
\newblock \href {http://arxiv.org/abs/2504.12516} {Browsecomp: A simple yet challenging benchmark for browsing agents}.

\bibitem[{Wei et~al.(2022)Wei, Wang, Schuurmans, Bosma, Xia, Chi, Le, Zhou et~al.}]{wei2022chain}
Jason Wei, Xuezhi Wang, Dale Schuurmans, Maarten Bosma, Fei Xia, Ed~Chi, Quoc~V Le, Denny Zhou, et~al. 2022.
\newblock Chain-of-thought prompting elicits reasoning in large language models.
\newblock \emph{Advances in neural information processing systems}, 35:24824--24837.

\bibitem[{Wolfson et~al.(2022)Wolfson, Deutch, and Berant}]{wolfson-etal-2022-weakly}
Tomer Wolfson, Daniel Deutch, and Jonathan Berant. 2022.
\newblock \href {https://doi.org/10.18653/v1/2022.findings-naacl.193} {Weakly supervised text-to-{SQL} parsing through question decomposition}.
\newblock In \emph{Findings of the Association for Computational Linguistics: NAACL 2022}, pages 2528--2542, Seattle, United States. Association for Computational Linguistics.

\bibitem[{Wolfson et~al.(2020)Wolfson, Geva, Gupta, Gardner, Goldberg, Deutch, and Berant}]{wolfson-etal-2020-break}
Tomer Wolfson, Mor Geva, Ankit Gupta, Matt Gardner, Yoav Goldberg, Daniel Deutch, and Jonathan Berant. 2020.
\newblock \href {https://doi.org/10.1162/tacl_a_00309} {Break it down: A question understanding benchmark}.
\newblock \emph{Transactions of the Association for Computational Linguistics}, 8:183--198.

\bibitem[{Wong et~al.(2025)Wong, Wang, Zhao, Chen, Gao, Zhang, Zhou, Wang, Xiang, Zhang, Huang, Wang, and Wang}]{wong2025widesearchbenchmarkingagenticbroad}
Ryan Wong, Jiawei Wang, Junjie Zhao, Li~Chen, Yan Gao, Long Zhang, Xuan Zhou, Zuo Wang, Kai Xiang, Ge~Zhang, Wenhao Huang, Yang Wang, and Ke~Wang. 2025.
\newblock \href {http://arxiv.org/abs/2508.07999} {Widesearch: Benchmarking agentic broad info-seeking}.

\bibitem[{Yang et~al.(2024)Yang, Sun, Xin, Sun, Bhalla, Chen, Choudhary, Gui, Jiang, JIANG, Kong, Moran, Wang, Xu, Yan, Yang, Yuan, Zha, Tang, Chen, SCHEFFER, Liu, Shah, Wanga, Kumar, tau Yih, and Dong}]{yang2024crag}
Xiao Yang, Kai Sun, Hao Xin, Yushi Sun, Nikita Bhalla, Xiangsen Chen, Sajal Choudhary, Rongze Gui, Ziran Jiang, Ziyu JIANG, Lingkun Kong, Brian Moran, Jiaqi Wang, Yifan~Ethan Xu, An~Yan, Chenyu Yang, Eting Yuan, Hanwen Zha, Nan Tang, Lei Chen, Nicolas SCHEFFER, Yue Liu, Nirav Shah, Rakesh Wanga, Anuj Kumar, Wen tau Yih, and Xin~Luna Dong. 2024.
\newblock \href {https://openreview.net/forum?id=Q7lAqY41HH} {{CRAG} - comprehensive {RAG} benchmark}.
\newblock In \emph{The Thirty-eight Conference on Neural Information Processing Systems Datasets and Benchmarks Track}.

\bibitem[{Yang et~al.(2018)Yang, Qi, Zhang, Bengio, Cohen, Salakhutdinov, and Manning}]{yang-etal-2018-hotpotqa}
Zhilin Yang, Peng Qi, Saizheng Zhang, Yoshua Bengio, William Cohen, Ruslan Salakhutdinov, and Christopher~D. Manning. 2018.
\newblock \href {https://doi.org/10.18653/v1/D18-1259} {{H}otpot{QA}: A dataset for diverse, explainable multi-hop question answering}.
\newblock In \emph{Proceedings of the 2018 Conference on Empirical Methods in Natural Language Processing}, pages 2369--2380, Brussels, Belgium. Association for Computational Linguistics.

\bibitem[{Yoran et~al.(2024{\natexlab{a}})Yoran, Amouyal, Malaviya, Bogin, Press, and Berant}]{yoran2024assistantbenchwebagentssolve}
Ori Yoran, Samuel~Joseph Amouyal, Chaitanya Malaviya, Ben Bogin, Ofir Press, and Jonathan Berant. 2024{\natexlab{a}}.
\newblock \href {http://arxiv.org/abs/2407.15711} {Assistantbench: Can web agents solve realistic and time-consuming tasks?}

\bibitem[{Yoran et~al.(2023)Yoran, Wolfson, Bogin, Katz, Deutch, and Berant}]{yoran-etal-2023-answering}
Ori Yoran, Tomer Wolfson, Ben Bogin, Uri Katz, Daniel Deutch, and Jonathan Berant. 2023.
\newblock \href {https://doi.org/10.18653/v1/2023.emnlp-main.364} {Answering questions by meta-reasoning over multiple chains of thought}.
\newblock In \emph{Proceedings of the 2023 Conference on Empirical Methods in Natural Language Processing}, pages 5942--5966, Singapore. Association for Computational Linguistics.

\bibitem[{Yoran et~al.(2024{\natexlab{b}})Yoran, Wolfson, Ram, and Berant}]{yoran2024making}
Ori Yoran, Tomer Wolfson, Ori Ram, and Jonathan Berant. 2024{\natexlab{b}}.
\newblock \href {https://openreview.net/forum?id=ZS4m74kZpH} {Making retrieval-augmented language models robust to irrelevant context}.
\newblock In \emph{The Twelfth International Conference on Learning Representations}.

\bibitem[{Yu et~al.(2018)Yu, Zhang, Yang, Yasunaga, Wang, Li, Ma, Li, Yao, Roman, Zhang, and Radev}]{yu-etal-2018-spider}
Tao Yu, Rui Zhang, Kai Yang, Michihiro Yasunaga, Dongxu Wang, Zifan Li, James Ma, Irene Li, Qingning Yao, Shanelle Roman, Zilin Zhang, and Dragomir Radev. 2018.
\newblock \href {https://doi.org/10.18653/v1/D18-1425} {{S}pider: A large-scale human-labeled dataset for complex and cross-domain semantic parsing and text-to-{SQL} task}.
\newblock In \emph{Proceedings of the 2018 Conference on Empirical Methods in Natural Language Processing}, pages 3911--3921, Brussels, Belgium. Association for Computational Linguistics.

\bibitem[{Zhang and Choi(2021)}]{zhang-choi-2021-situatedqa}
Michael Zhang and Eunsol Choi. 2021.
\newblock \href {https://doi.org/10.18653/v1/2021.emnlp-main.586} {{S}ituated{QA}: Incorporating extra-linguistic contexts into {QA}}.
\newblock In \emph{Proceedings of the 2021 Conference on Empirical Methods in Natural Language Processing}, pages 7371--7387, Online and Punta Cana, Dominican Republic. Association for Computational Linguistics.

\bibitem[{Zhong et~al.(2023)Zhong, Shi, Yih, and Zettlemoyer}]{zhong-etal-2023-romqa}
Victor Zhong, Weijia Shi, Wen-tau Yih, and Luke Zettlemoyer. 2023.
\newblock \href {https://doi.org/10.18653/v1/2023.findings-emnlp.470} {{R}o{MQA}: A benchmark for robust, multi-evidence, multi-answer question answering}.
\newblock In \emph{Findings of the Association for Computational Linguistics: EMNLP 2023}, pages 7055--7067, Singapore. Association for Computational Linguistics.

\bibitem[{Zhu et~al.(2024)Zhu, Hwang, Dugan, and Callison-Burch}]{zhu2024fanoutqamultihopmultidocumentquestion}
Andrew Zhu, Alyssa Hwang, Liam Dugan, and Chris Callison-Burch. 2024.
\newblock \href {http://arxiv.org/abs/2402.14116} {Fanoutqa: A multi-hop, multi-document question answering benchmark for large language models}.

\end{thebibliography}
